\documentclass[10pt,twocolumn,letterpaper]{article}

\usepackage[pagenumbers]{cvpr} 

\usepackage{graphicx}
\usepackage{amsmath}
\usepackage{amssymb}
\usepackage{booktabs}
\usepackage{multirow}
\usepackage{enumitem}
\usepackage{appendix}

\usepackage[pagebackref,breaklinks,colorlinks]{hyperref}
\usepackage[linesnumbered,ruled,vlined]{algorithm2e}

\usepackage[capitalize]{cleveref}
\crefname{section}{Sec.}{Secs.}
\Crefname{section}{Section}{Sections}
\Crefname{table}{Table}{Tables}
\crefname{table}{Tab.}{Tabs.}

\newcommand{\squeezeup}{\vspace{-4mm}}
\newcommand{\vect}[1]{\boldsymbol{#1}}
\newcommand{\squeezeupsmall}{\vspace{-2mm}}

\def\cvprPaperID{6387} 
\def\confName{CVPR}
\def\confYear{2023}

\begin{document}
\title{TrustMark: Universal Watermarking for Arbitrary Resolution Images}
\author{Tu Bui$^{1}$ \\
\and Shruti Agarwal$^{2}$ \\
\and John Collomosse$^{1,2}$\\
\and \vspace{-1.1cm}\\
$^{1}$University of Surrey, $^{2}$Adobe Research\\
{\tt\small t.v.bui@surrey.ac.uk, \{shragarw,collomos\}@adobe.com}}

\maketitle

\begin{abstract}
   Imperceptible digital watermarking is important in copyright protection, misinformation prevention, and responsible generative AI. We propose TrustMark - a GAN-based watermarking method with novel design in architecture and spatio-spectra losses to balance the trade-off between watermarked image quality with the watermark recovery accuracy. Our model is trained with robustness in mind, withstanding various in- and out-place perturbations on the encoded image. Additionally, we introduce TrustMark-RM - a watermark remover method useful for re-watermarking. Our methods achieve state-of-art performance on 3 benchmarks comprising arbitrary resolution images.
\end{abstract}
\squeezeup
\section{Introduction}
\label{sec:intro}

Recent advances in generative AI (GenAI) present fresh challenges in combating misinformation, safeguarding copyright, and identifying the origins of content.  Emerging content provenance standards \cite{c2pa} embed signed metadata within assets to describe how an image was created, by whom, and by what means.  However, such metadata is often stripped as assets are re-shared and distributed online.  Digital watermarking offers a way to embed an imperceptible identifier that may be used to recover provenance information in this situation.


The goal of (impercetible) image watermarking is to embed a secret (\eg provenance data) directly within the image content in such a way that the changes to the image are invisible to the naked eye, yet detectable by a `watermark decoder'. Since the image size does not change, certain image content has to be sacrificed to accommodate the watermark. Early statistical methods leverage digital computing theory to identify the lowest entropy regions \eg least significant bit in pixel or frequency domain, regardless of the image content. More recent approaches adopt deep learning for content-aware watermark embedding. While imperceptibility has been steadily improved, the robustness aspects have been largely overlooked -- often only few common noise types such as jpeg compression, Gaussian blur and cropping are taken into account \cite{zhu2018hidden,zhang2019robust,tancik2020stegastamp,tavakoli2023convolutional,navas2008dwt}. Moreover, many robust methods work only over fixed resolutions of $\sim 200$ pixels \cite{zhu2018hidden,tancik2020stegastamp,luo2020distortion} -- an order of magnitude smaller than images typically used for creative work.   Another key consideration for creative practice is quality degradation due to watermarking, and `re-watermarking' (watermark removal and replacement) as images are processed by creative tools.   

We propose TrustMark, a novel watermarking method to address these challenges. Our contributions are four-fold:\\
    \textbf{1. Novel architecture} with $1 \times 1$ convolutions in the post-process layers and focal frequency loss for improved preservation of high frequency detail in the watermarked image.  Robustness of the encoding is encouraged via extensive noise simulation during training.\\
    \textbf{2. Resolution scaling} method to extend TrustMark and other watermarking methods to operate over images of arbitrary resolution.\\
    \textbf{3. Watermark removal network} TrustMark-RM, to restore the original image with high quality, useful for applications such as re-watermarking.\\
    \textbf{4. State-of-the-art performance} in both imperceptibility and watermark recovery on three benchmarks.

\section{Related work}
\label{sec:relate}

\begin{figure*}[]
    \centering
    \includegraphics[width=1.0\linewidth,trim=0cm 0cm 0cm 0cm,clip,height=8cm]{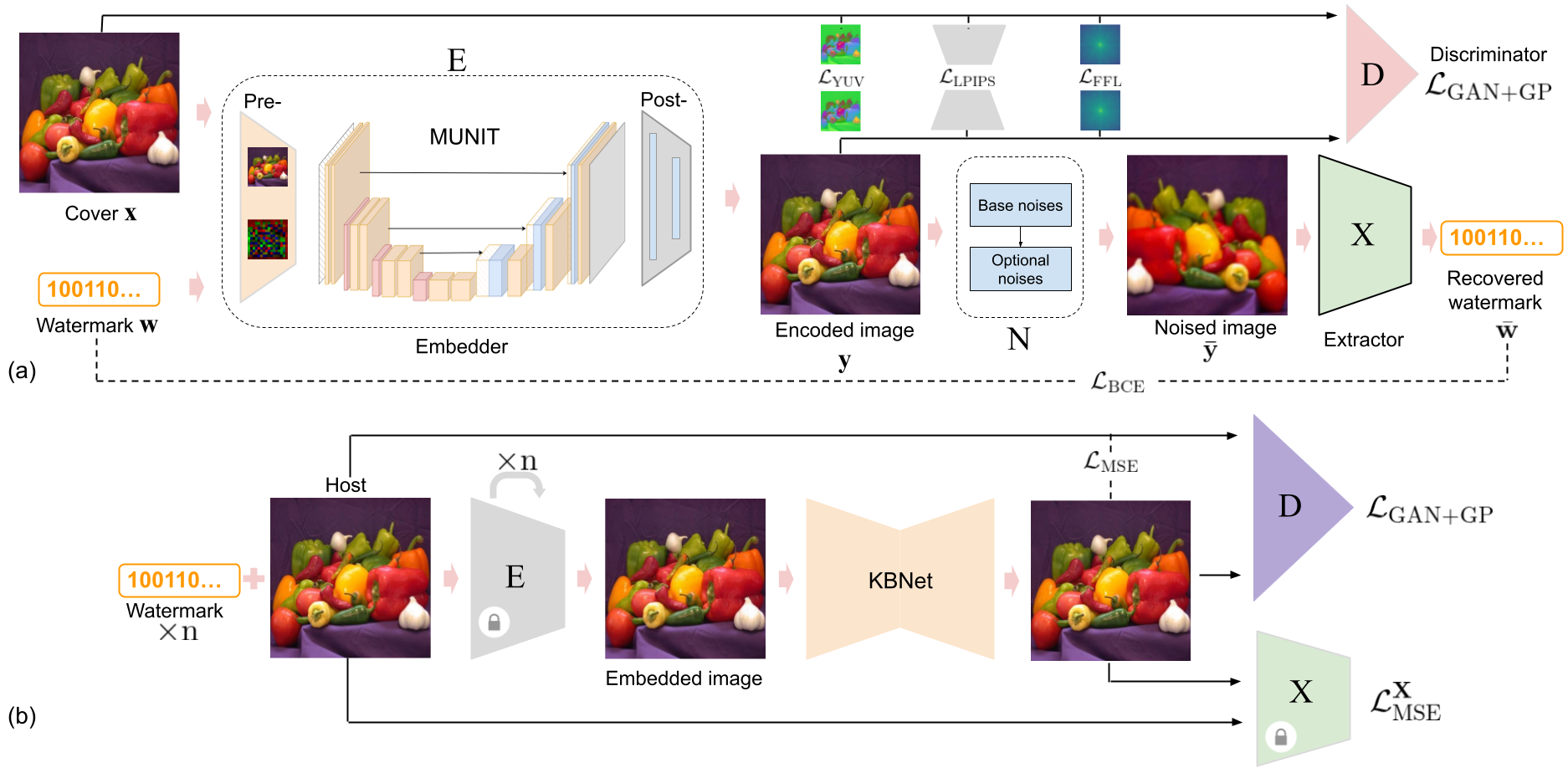}
    \squeezeup
    \caption{Proposed architecture of TrustMark (a). The embedder \textbf{E} encodes a watermark into a cover image robustly using   
     a noise module \textbf{N} to simulate common perturbations on the encoded image.  The extractor \textbf{X} recovers the watermark from the encoded image.  The TrustMark-RM network (b) removes the watermark to enable re-watermarking of the image.}
    \squeezeup
    \label{fig:arch}
\end{figure*}

\noindent\textbf{Media provenance} is the focus of cross-industry coalitions (\eg CAI \cite{cai}, Origin \cite{origin}) and emerging standards (\eg C2PA \cite{c2pa}) that encode a metadata `manifest' within an image.  When metadata is stripped, a perceptual hash \cite{nguyen2021,Bharati2021,Black_2021_CVPR} can be used to lookup the manifest in a database \cite{origin} or distributed ledger \eg blockchain \cite{arch2,arch3}.  However reliance on near-duplicate search is inexact, and provides no signal to trigger such a lookup. (Re-)watermarking provides an alternative way -- to insert and update an imperceptible identifer to recover stripped metadata.

\noindent\textbf{Classical Watermarking} explored Least Significant Bit (LSB)~\cite{wolfgang1996watermark} to embed a secret in the lowest order bits of each pixel, producing images  perceptually indistinguishable from the original (`cover') image. Since then, several techniques have exploited the spatial~\cite{taha2022high, ghazanfari2011lsb++} and frequency~\cite{navas2008dwt, li2007steganographic, pevny2010using, holub2012designing, holub2014universal,outguess} domains to embed the secret. Although such methods can embed large payloads imperceptibly, they suffer with poor robustness to even minor modifications to the encoded image. 

\noindent\textbf{Deep Watermarking} have been shown to provide better robustness to noises while maintaining good quality of the generated image~\cite{wan2022comprehensive}. HiDDeN~\cite{zhu2018hidden} was the first end-to-end trained watermarking network that used the encoder-decoder architecture for watermark embedding. However, HiDDeN embeds the watermark at every pixel location therefore does not scale well with the watermark size.  Using the similar encoder-decoder approach, a variety of architectures were proposed for the improvement of stego image quality and robustness in several later works~\cite{tancik2020stegastamp,zhang2019robust,chang2021neural,duan2019reversible,wu2018stegnet,meng2018fusion}. These works mostly encode secrets and covers jointly, often with an UNet-like model and skip connections to preserve small details in the cover images \cite{tancik2020stegastamp,duan2019reversible,zhang2019robust}. Notably, StegaStamp \cite{tancik2020stegastamp} incorporates spatial transformers for robustness against geometric transformations. RivaGAN \cite{zhang2019robust} employs attention mechanism for video watermarking. SSL \cite{fernandez2022watermarking} takes a different approach, watermarking images in the latent space at inference time via back-propagation, achieving superior imperceptibility score at cost of speed. Recently, RoSteALS \cite{bui2023rosteals} also proposes to watermark via the latent code of a frozen VQVAE~\cite{esser2021taming}, achieving state-of-art robustness however its imperceptibility is limited by VQVAE~\cite{esser2021taming} reconstruction quality.

\noindent\textbf{Watermark removal} has been investigated in the context of visible watermark removal for either adversarial watermark removal~\cite{cao2019generative,niu2023fine,cun2021split,liu2021wdnet} or for improving the watermark robustness against adversarial attacks~\cite{chen2023universal,liu2022watermark,lyu2023adversarial}. In this paper, we investigate removal of invisible watermarks for the task of image re-encryption. 


\begin{figure*}[]
    \centering
    \includegraphics[width=1.0\linewidth,trim=0cm 0cm 0cm 0cm,clip,height=11cm]{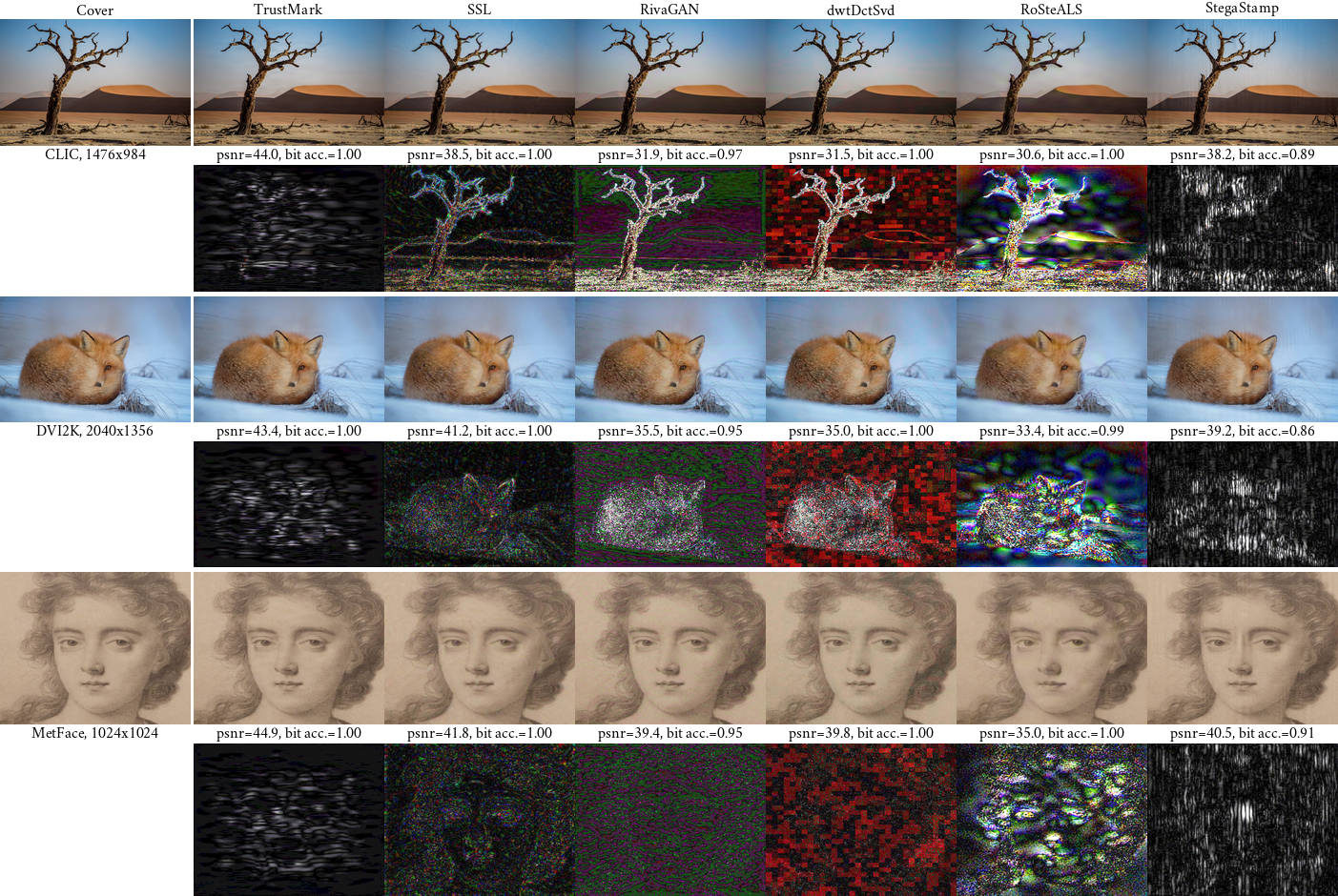}
    \squeezeup
    \caption{Representative examples of TrustMark and 5 baseline methods (SSL \cite{fernandez2022watermarking}, RivaGAN \cite{zhang2019robust}, dwtDctSvd \cite{navas2008dwt}, RoSteALS~\cite{bui2023rosteals}, StegaStamp \cite{tancik2020stegastamp}) over 3 benchmarks (CLIC \cite{clic2020}, DIV2k \cite{div2k}) and MetFace \cite{metface}).  The residual is amplified $20 \times$ for visualization.}
    \squeezeup
    \label{fig:eg}
\end{figure*}
\section{Methodology}
\label{sec:method}
We treat watermarking and removal as 2 separate problems, since the former involves trading-off between two data sources (watermark and image content) while the latter resembles a denoising task. We describe the watermarking network (TrustMark) and the watermark removal network (TrustMark-RM) in \cref{sec:watermark} and \cref{sec:wm_removal} respectively; and outline how the two can be adapted to work on arbitrary resolution images at inference time in \cref{sec:arbitrary_res}. 

\subsection{Watermarking network}
\label{sec:watermark}
The TrustMark architecture (\cref{fig:arch}a) comprises an embedder module \textbf{E} to encode the watermark into a host (`cover') image, an extractor module \textbf{X} to recover the watermark from that image, and a noise module \textbf{N} to perturb the image during training. We  describe each in turn.
\subsubsection{Watermark Embedder}
The embedder network \textbf{E} first accepts a cover image \textbf{x} $\in \mathbb{R}^{256\times 256\times 3}$ and a watermark $\vect{w}$ $\in \{0,1\}^{l}$ (with $l$ being the watermark size) into its pre-processing module $\textbf{E}_{\mathrm{pre}}(\textbf{x},\vect{w}) \in \mathbb{R}^{256\times 256\times d}$, where $d$ is the internal feature dimension. Following \cite{tancik2020stegastamp,baluja2017hiding}, we design $\textbf{E}_{\mathrm{pre}}$ as an early image-watermark fusion network -- the watermark is interpolated to match the cover image's dimension, then the concatenated cover-watermark feature map is convolved with $d$ $3\times3$ filters to make a $d$-channel image output. Next, we employ a MUNIT-based \cite{munit} network customized for watermark embedding. Specifically, we remove channel normalization, double the network depth but halve the internal dimension, effectively reducing the model size 2x. 
Finally, a post-process module converts MUNIT's n-channel output back to the RGB space, $\bar{\textbf{x}}=\textbf{E}_{\mathrm{post}}(.) \in \mathbb{R}^{256\times 256\times 3}$. While this could be implemented using just a single convolutional layer \cite{tancik2020stegastamp,baluja2017hiding,zhu2018hidden}, a more complex $\textbf{E}_{\mathrm{post}}$ is needed to retain fidelty in the encoded image, especially high frequency details. We therefore leverage multiple $1 \times 1$ convolution layers which act as channel-wise pooling layers commonly used in dimensionality reduction and feature learning networks \cite{netinnet,inceptionv1}. These $1 \times 1$ convolution layers are separated by SiLU and end with a $tanh()$ activation to constraint the output pixel values to range $[-1,1]$. Note that our embeder \textbf{E} outputs the encoded image directly, $\textbf{y}=\textbf{E}(\textbf{x},\vect{w})$, instead of estimating the residual artifacts to be added to the cover image as in StegaStamp \cite{tancik2020stegastamp} or RivaGAN \cite{zhang2019robust}.     

\subsubsection{Watermark extractor}
The extractor network \textbf{X} aims to decode the watermark from the encoded image, $\bar{\vect{w}}=\textbf{X}(\textbf{y}) \in \{0,1\}^l$. This is challenging as the watermark signal is perceptually invisible. We observe that only resnet-based networks supervised under a particular training scheme fit the task (c.f. \cref{sec:data} and \cref{sec:ablation}).  Here we employ the standard ResNet50 \cite{resnet} with the last layer being replaced by a $l$-dimension sigmoid-activated FC to predict the $l$-bit watermark.
\vspace{-2mm}
\subsubsection{Noise model}
\label{sec:noise}
Robustness to noises is an important factor for invisible watermarking. To expose the extractor \textbf{X} to various noise sources when it is being jointly trained with the embedder \textbf{E}, we insert a noise model \textbf{N} after the encoded image - $\Tilde{\textbf{y}}=\textbf{N}(\textbf{y})$. \textbf{N} consists of 3 geometrical transformations (random flip, crop and resize) and 15 perturbation sources (random JPEG compression, brightness, hue, contrast, sharpness, color jitter, RGB shift, saturation, grayscale, Gaussian blur, median blur, box blur, motion blur, Gaussian noise, posterize). Each encoded image is perturbed with 3 geometrical transformations (`base transforms' in \cref{fig:arch}) and 2 other random noises (`optional transforms'). All 18 transforms are differentiable so that errors can be propagated back to the embedder.  More details are in the Sup.Mat.

\begin{algorithm}[ht]
\DontPrintSemicolon
  \newcommand\mycommfont[1]{\footnotesize\ttfamily\textcolor{blue}{#1}}
  \SetCommentSty{mycommfont}
  \SetKwInput{KwInput}{Input}                
  \SetKwInput{KwOutput}{Output}              
  \KwInput{Input image \textbf{x}, [binary watermark vector $\vect{w}$]}
  \KwOutput{Restored image \textbf{y}}
  \KwData{Embedding network \textbf{E} or Removal network \textbf{R}}

  H, W := \textbf{x}.height, \textbf{x}.width

  $\textbf{x} \gets \textbf{x} /127.5 - 1$ \tcp*{Normalize to range [-1,1]}

  $\bar{\textbf{x}} := \mathrm{interpolate}(\textbf{x}, (256,256))$
  
  \If {model is watermarking}
  {
  $\textbf{r} := \textbf{E}(\bar{\textbf{x}}, \vect{w}) - \bar{\textbf{x}}$ \tcp*{residual image}
  }
  \Else{
  $\textbf{\textbf{r}} := R(\bar{\textbf{x}}) - \bar{\textbf{x}}$
  }
  $\textbf{r} \gets \mathrm{interpolate}(\textbf{r}, (\text{H},\text{W})) $

  $\textbf{y} \gets \mathrm{clamp}(\textbf{x} + \textbf{r}, -1,1)$

  $\textbf{y} \gets \textbf{y}*127.5+127.5$

\caption{Resolution scaling - watermark embedding or removal on arbitrary resolution images.}
\label{alg:arbitrary}
\end{algorithm}

\begin{table*}[t!]
\centering
\definecolor{Gray}{gray}{0.9}
\small
\begin{tabular}{l|cc|cc|cc}
\toprule
\multirow{2}{*}{Method} & \multicolumn{2}{c}{CLIC} & \multicolumn{2}{c}{DIV2K} & \multicolumn{2}{c}{MetFace} \\
 & PSNR & SSIM & PSNR & SSIM & PSNR & SSIM \\
\midrule
TrustMark-Q ($\alpha_{\mathrm{max}}=27.5$) & \textbf{43.26$\pm$1.59} & \textbf{0.99$\pm$0.00} & 42.39$\pm$1.46 & \textbf{0.99$\pm$0.00} & \textbf{45.34$\pm$1.33} & \textbf{0.99$\pm$0.00} \\
TrustMark-B ($\alpha_{\mathrm{max}}=20$) & 41.53$\pm$1.84 & \textbf{0.99$\pm$0.01} & 40.20$\pm$1.87 & \textbf{0.99$\pm$0.01} & 43.87$\pm$1.42 & \textbf{0.99$\pm$0.00}  \\
RoSteALS~\cite{bui2023rosteals} & 30.03$\pm$2.63 & 0.92$\pm$0.04 & 27.95$\pm$2.51 & 0.88$\pm$0.05 & 33.77$\pm$2.37 & 0.93$\pm$0.03\\
RivaGAN \cite{zhang2019robust} & 41.04$\pm$0.31&0.98$\pm$0.01 & 41.06$\pm$0.35&\textbf{0.99$\pm$0.01} & 40.98$\pm$0.19&0.98$\pm$0.01 \\
SSL \cite{fernandez2022watermarking} & 42.74$\pm$0.12& \textbf{0.99$\pm$0.01} & \textbf{42.73$\pm$0.12}&\textbf{0.99$\pm$0.01} & 42.84$\pm$0.10& \textbf{0.99$\pm$0.00} \\
StegaStamp \cite{tancik2020stegastamp} & 37.48$\pm$1.93 & \textbf{0.99$\pm$0.02} & 35.87$\pm$1.73 & 0.98$\pm$0.02  & 39.35$\pm$1.57 & \textbf{0.99$\pm$0.01} \\
dwtDctSvd \cite{navas2008dwt} & 39.13$\pm$1.21&0.98$\pm$0.01 & 38.02$\pm$1.35&0.97$\pm$0.02 & 41.14$\pm$2.35&0.98$\pm$0.01\\
\midrule
 & Acc. (clean) & Acc. (noised) & Acc. (clean) & Acc. (noised) & Acc. (clean) & Acc. (noised) \\
\midrule
TrustMark-Q ($\alpha_{\mathrm{max}}=27.5$) & \textbf{1.00} & 0.95$\pm$0.09 & \textbf{1.00} & 0.95$\pm$0.09 & \textbf{1.00} & 0.96$\pm$0.10 \\
TrustMark-B ($\alpha_{\mathrm{max}}=20$) & \textbf{1.00} & \textbf{0.97$\pm$0.08} & 0.99$\pm$0.01 & \textbf{0.97$\pm$0.06} & \textbf{1.00} & \textbf{0.97$\pm$0.07} \\
RoSteALS~\cite{bui2023rosteals} & 0.99$\pm$0.02 & 0.94$\pm$0.09 & 0.99$\pm$0.02 & 0.93$\pm$0.09 & 0.99$\pm$0.01 & 0.93$\pm$0.10 \\

RivaGAN \cite{zhang2019robust} & 0.96$\pm$0.04 & 0.79$\pm$0.14 & 0.96$\pm$0.05 & 0.78$\pm$0.14 & 0.97$\pm$0.02 & 0.82$\pm$0.14 \\

SSL \cite{fernandez2022watermarking} & 0.79$\pm$0.09 & 0.60$\pm$0.09 & 0.70$\pm$0.09 & 0.57$\pm$0.07 & \textbf{1.00} & 0.70$\pm$0.13\\

StegaStamp \cite{tancik2020stegastamp} & 0.88$\pm$0.04 & 0.72$\pm$0.10 & 0.85$\pm$0.05 & 0.70$\pm$0.10 & 0.89$\pm$0.03 & 0.72$\pm$0.11 \\

dwtDctSvd \cite{navas2008dwt}& \textbf{1.00} & 0.52$\pm$0.06 & 0.99$\pm$0.05 & 0.51$\pm$0.06 & \textbf{1.00} & 0.52$\pm$0.08\\
\bottomrule
\end{tabular}
\caption{TrustMark versus baselines on three benchmarks. Imperceptibility is measured using PSNR and SSIM metrics, while watermark recovery is evaluated via bit accuracy on the watermarked images before (clean) and after (noised) it is exposed to random noises.}
\label{tab:main}
\end{table*}
\squeezeup
\subsubsection{Losses}
Overall, training TrustMark involves balancing image quality (via \textbf{E}) with watermark recovery (via \textbf{X}) in the presence of complex noise simulation between. 

\begin{equation}
    \mathcal{L}_{total} = \alpha \mathcal{L}_{\mathrm{quality}}(\textbf{x}, \textbf{y}) + \mathcal{L}_{\mathrm{recovery}}(\vect{w}, \bar{\vect{w}})
    \label{eq:loss_all}
\end{equation}
where $\alpha$ is the trade-off hyper-parameter. We adopt RoSteALS \cite{bui2023rosteals} strategy to start training with a low value of $\alpha$ to prioritize watermark recovery then linearly increase to a threshold $\alpha_{\mathrm{max}}$, which can be set prior training to exert TrustMark's controlability (see \cref{exp:emb}).

$\mathcal{L}_{\mathrm{recovery}}(\vect{w}, \bar{\vect{w}})$ is the standard binary cross-entropy loss to bring the recovered watermark close to the original. The quality loss is defined as,
\begin{align}
    \mathcal{L}_{\mathrm{quality}}(\textbf{x}, \textbf{y}) =& \beta_{\mathrm{YUV}} \mathcal{L}_{\mathrm{YUV}} + \beta_{\mathrm{LPIPS}}\mathcal{L}_{\mathrm{LPIPS}} \\
    &+\beta_{\mathrm{FFL}} \mathcal{L}_{\mathrm{FFL}} + \beta_{\mathrm{GAN}}\mathcal{L}_{\mathrm{GAN}+\mathrm{GP}}
\end{align}

where $\beta_{\mathrm{YUV}}, \beta_{\mathrm{LPIPS}}, \beta_{\mathrm{FFL}}$ and $\beta_{\mathrm{GAN}}$ are the weights of 4 loss terms. $\mathcal{L}_{\mathrm{YUV}}(\textbf{x},\textbf{y})$ is the mean squared error loss in the YUV pixel space, $\mathcal{L}_{\mathrm{LPIPS}}(\textbf{x},\textbf{y})$ is the perceptual loss following \cite{bui2023rosteals,tancik2020stegastamp}. Additionally, TrustMark is also trained in generative adversarial fashion with GAN loss,
\begin{align}
    \mathcal{L}_{\mathrm{GAN}+\mathrm{GP}}(\textbf{x},\textbf{y}) =& \underset{\textbf{y} \sim \mathbb{P}_E}{\mathbb{E}}[\textbf{D}(\textbf{y})] - \underset{\textbf{x} \sim \mathbb{P}_{real}}{\mathbb{E}}[\textbf{D}(\textbf{x})] \\
    &+\lambda \underset{\textbf{y} \sim \mathbb{P}_E}{\mathbb{E}}[(\left|\left| \nabla_y \textbf{D}(\textbf{y}) \right|\right|_2-1)^2]
\end{align}
where \textbf{D} is a discriminator to distinguish the encoded image from the original, $\lambda$ is the gradient penalty loss weight for training stablization \cite{wgangp}. 

Finally, we propose to add a focal frequency loss (FFL) to bridge the gap between the cover and encoded image in frequency domain,
\begin{equation}
    \mathcal{L}_{\mathrm{FFL}}(\textbf{x},\textbf{y}) = \rho_{f(\textbf{x}),f(\textbf{y})}\left|\left|f(\textbf{y})-f(\textbf{x}) \right|\right|_2
\end{equation}
where $f(.)$ is the 2D Fourier transform function and $\rho_{f(\textbf{x}),f(\textbf{y})} \in \mathbb{R}^{256\times256\times3}$ is a dynamic weight matrix to balance the loss across frequency spectrum. It is reported in \cite{dct,repmix} that synthezied images often exhibit artifacts in the frequency domain that can be easily spot by deep networks. FFL was first introduced in \cite{ffl2,ffl} to reduce such artifacts. Here we leverage FFL to encourage higher watermark embedding quality instead (see \cref{sec:ablation} and Sup.Mat).   


\subsection{Watermark removal}
\label{sec:wm_removal}
Watermark removal is the invert process of watermark embedding where the watermark artifacts are removed to reconstruct the cover image close to original image quality. Our proposed network, TrustMark-RM, is based on KBNet \cite{kbnet} -- the current state-of-art model for image restoration. KBNet leverages learnable kernel basis attention with multi-axis feature fusion, effectively combining adaptive spatial attention of transformers with the inductive bias strengths of convolution into one network design. TrustMark-RM's training scheme is depicted in \cref{fig:arch}(b), with the frozen watermark embedder $\textbf{E}(.)$ and extractor $\textbf{X}(.)$ are borrowed from TrustMark model. For each training cover image, we embed n random watermarks using $\textbf{E}(.)$ to synthesize the input sample to KBNet. The output of KBNet is regulated to be close to the cover image using 3 losses: (i) a pixel loss $\mathcal{L}_{\mathrm{MSE}}$ prioritizing Peak-Signal-To-Noise (PSNR) ratio directly, (ii) a discriminator loss $\mathcal{L}_{\mathrm{GAN}+\mathrm{GP}}$ similar to TrustMark, and (iii) watermark similarity loss $\mathcal{L}^{\textbf{X}}_{\mathrm{MSE}}$ to have the output image the same response to $\textbf{X}(.)$ as the cover (close to random chance).   

\subsection{Resolution scaling}
\label{sec:arbitrary_res}

TrustMark and TrustMark-RM accept an input image at fixed resolution of $256\times256$, however in-the-wild images often have variable resolution, often much higher than a model's receptive field. It is crucial to support watermarking and removal at the original resolution to retain the input image quality. We propose a simple yet effective method to adapt our models for arbitrary resolution in \cref{alg:arbitrary}. Note that our method offers 2 advantages: (i) it executes at inference time and treats the model as a blackbox, therefore can be applied for any watermarking/removal algorithms; and (ii) by computing and interpolating the residual artifacts, the high resolution output image can be derived directly from the original input (see L\#9 of \cref{alg:arbitrary}). This method requires the watermarking/removal models to be robust against resizing, which is addressed by TrustMark in \cref{sec:noise}. For TrustMark-RM, we show that artifact removal at the $256\times256$ resolution also retains its effects after Resolution Scaling is applied (\cref{sec:exp_wr}).     

\section{Experiments}
\label{sec:exp}

\subsection{Datasets, training details and baselines}
\label{sec:data}

\noindent \textbf{Datasets}. We follow the same settings of \cite{bui2023rosteals} to train our models on 101K images from the MIRFlickR 1M dataset \cite{mirflick} (1K images are used for validation) and evaluate on the CLIC \cite{clic2020} and MetFace \cite{metface} benchmarks. Additionally, we also evaluate on DIV2K \cite{div2k} -- a popular super-resolution benchmark consisting of 900 high quality train and validation images at 2K resolution. We use both the training and validation images of DIV2K for testing (the test set is not visible to public therefore not included). We note DIV2K is much diverse in content and has higher quality than the other 2 benchmarks. At test time, every image is associated with a random watermark and the encoded image is perturbed with random noises described in \cref{sec:noise}.

\begin{table}
    \centering
    \begin{tabular}{l|cc}
         \toprule
         Method&  PSNR& Acc. (noised)\\
         \midrule
         TrustMark ($\alpha_{\mathrm{max}}=15$) & 38.87$\pm$1.42  & 0.95$\pm$0.08\\
         RoSteALS~\cite{bui2023rosteals} &  32.68 $\pm$ 1.75 & 0.94 $\pm$ 0.07\\
         StegaStamp \cite{tancik2020stegastamp} & 31.26 $\pm$ 0.85 & 0.88 $\pm$ 0.13 \\
         SSL \cite{fernandez2022watermarking} & 41.84 $\pm$ 0.10 &0.62 $\pm$ 0.14 \\
         RivaGAN \cite{zhang2019robust} &  40.32 $\pm$ 0.15   &  0.77 $\pm$ 0.16\\
         dwtDctSvd  \cite{navas2008dwt} & 38.96 $\pm$ 1.41    & 0.61 $\pm$ 0.20\\
         \bottomrule
    \end{tabular}
    \caption{TrustMark versus baselines on CLIC dataset, using ImageNet-C noise configuration in training and evaluation. Baseline results are taken directly from \cite{bui2023rosteals}.}
    \label{tab:imgc}
\end{table}

\noindent \textbf{Training details}. We train TrustMark for 150 epochs with AdamW optimizer at initial learning rate $4e-6$ per image in a batch of 32s and cosine annealing schedule. Loss terms $\beta_{\mathrm{LPIPS}}, \beta_{\mathrm{YUV}}, \beta_{\mathrm{FFL}}$ and $\beta_{\mathrm{GAN}}$ are set to 1, 1.5, 1.5 and 1 respectively. It takes 48 hours to train on a Geforce RTX 3090 GPU and a standard Intel i7 processor.   For TrustMark-RM, we set n=3 and the training time is approx. 2 weeks for 100 epochs at batch size of 8 on an A100 GPU.   For inference, average watermarking encoding/decoding takes 125/25 milliseconds on a Nvidia RTX 3090 GPU.

\begin{figure}[]
    \centering
    \includegraphics[width=0.9\linewidth]{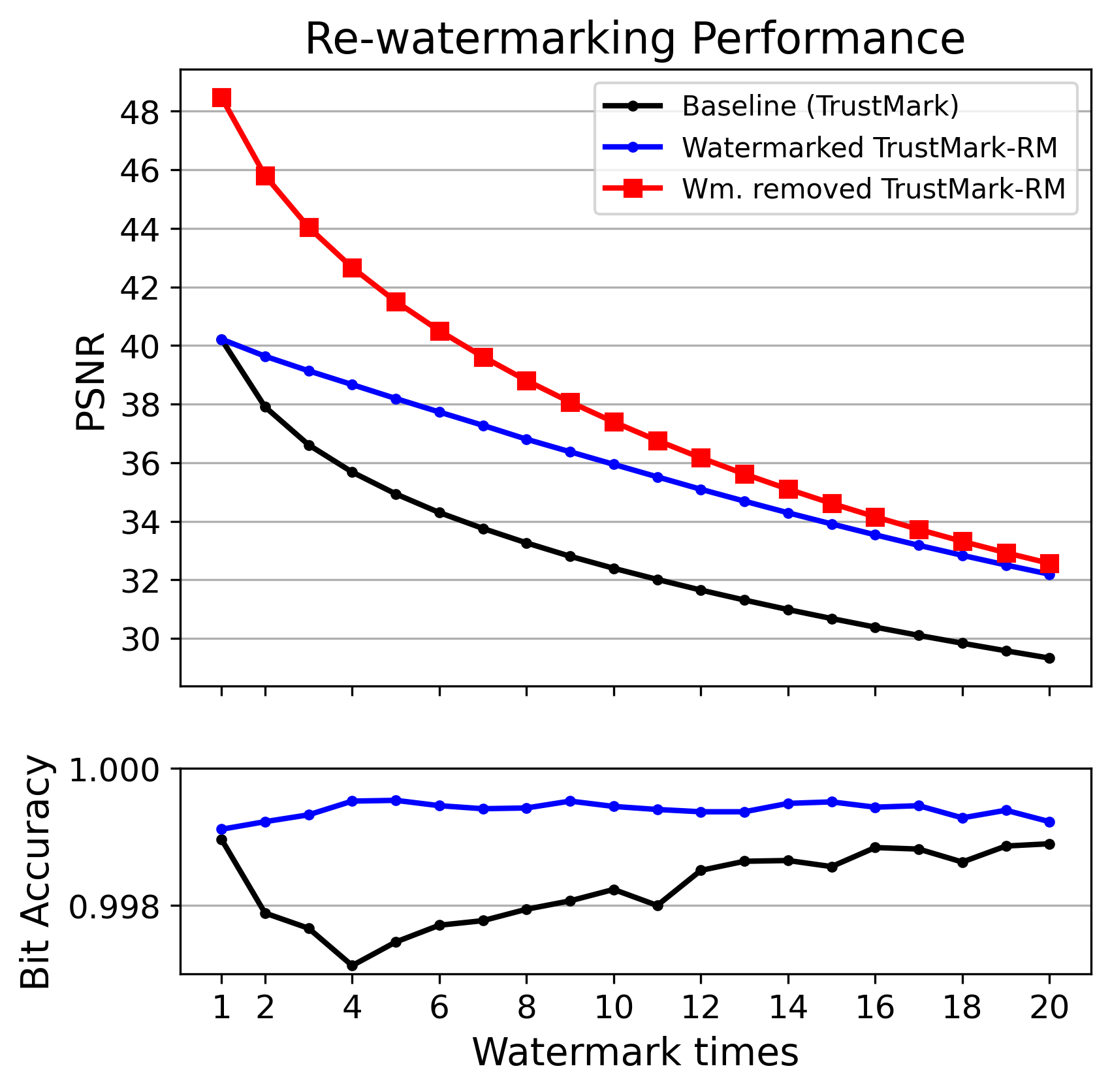}
    \squeezeup
    \caption{Re-watermarking performance with and without watermark removal, evaluated on the DIV2K dataset.}
    \squeezeup
    \label{fig:rewm}
\end{figure}

\begin{figure*}[]
    \centering
    \begin{tabular}{ccc}
    \includegraphics[width=0.3\linewidth,trim=0cm 0cm 0cm 0cm,clip]{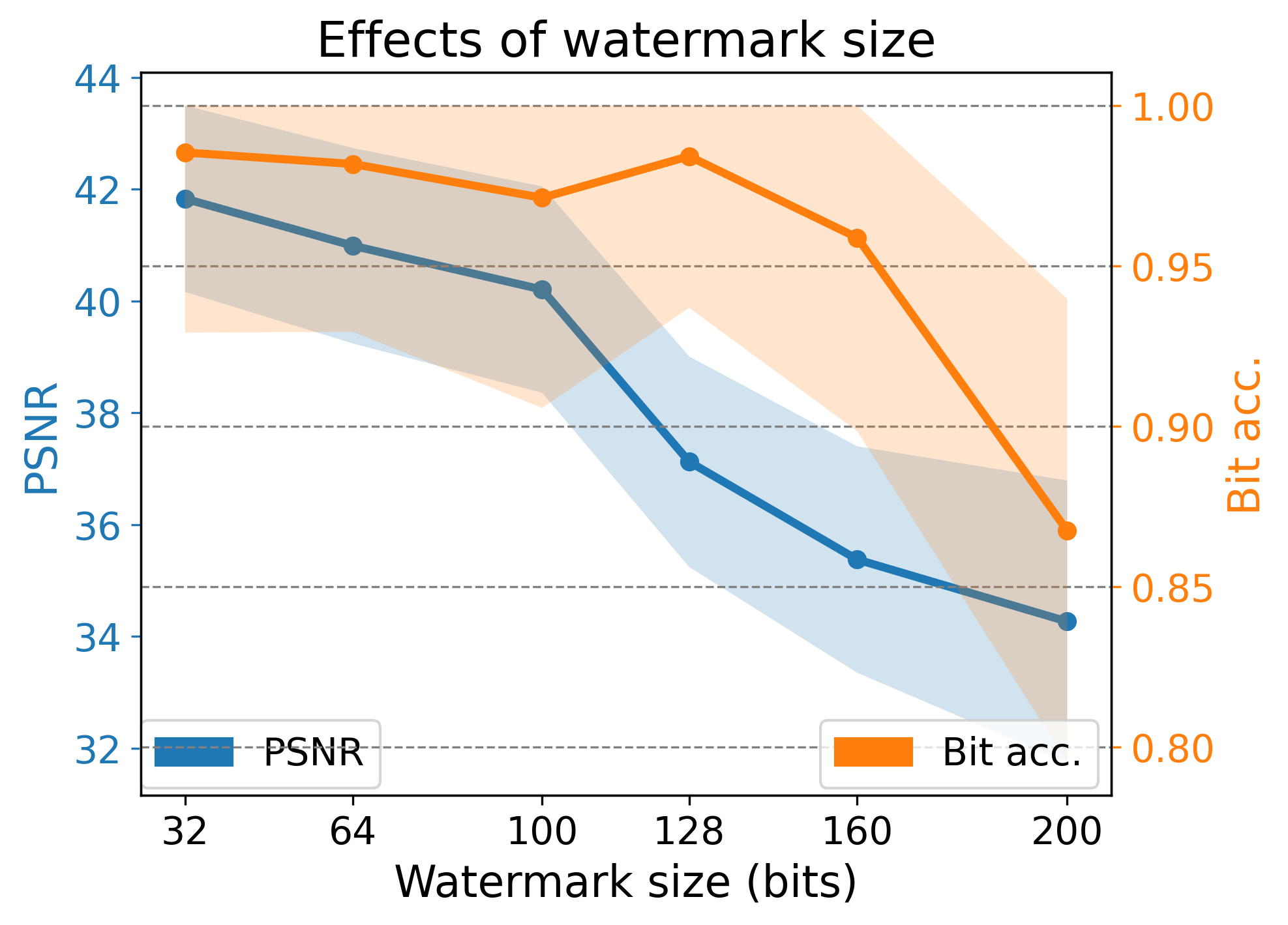} & 
    \includegraphics[width=0.3\linewidth,trim=0cm 0cm 0cm 0cm,clip]{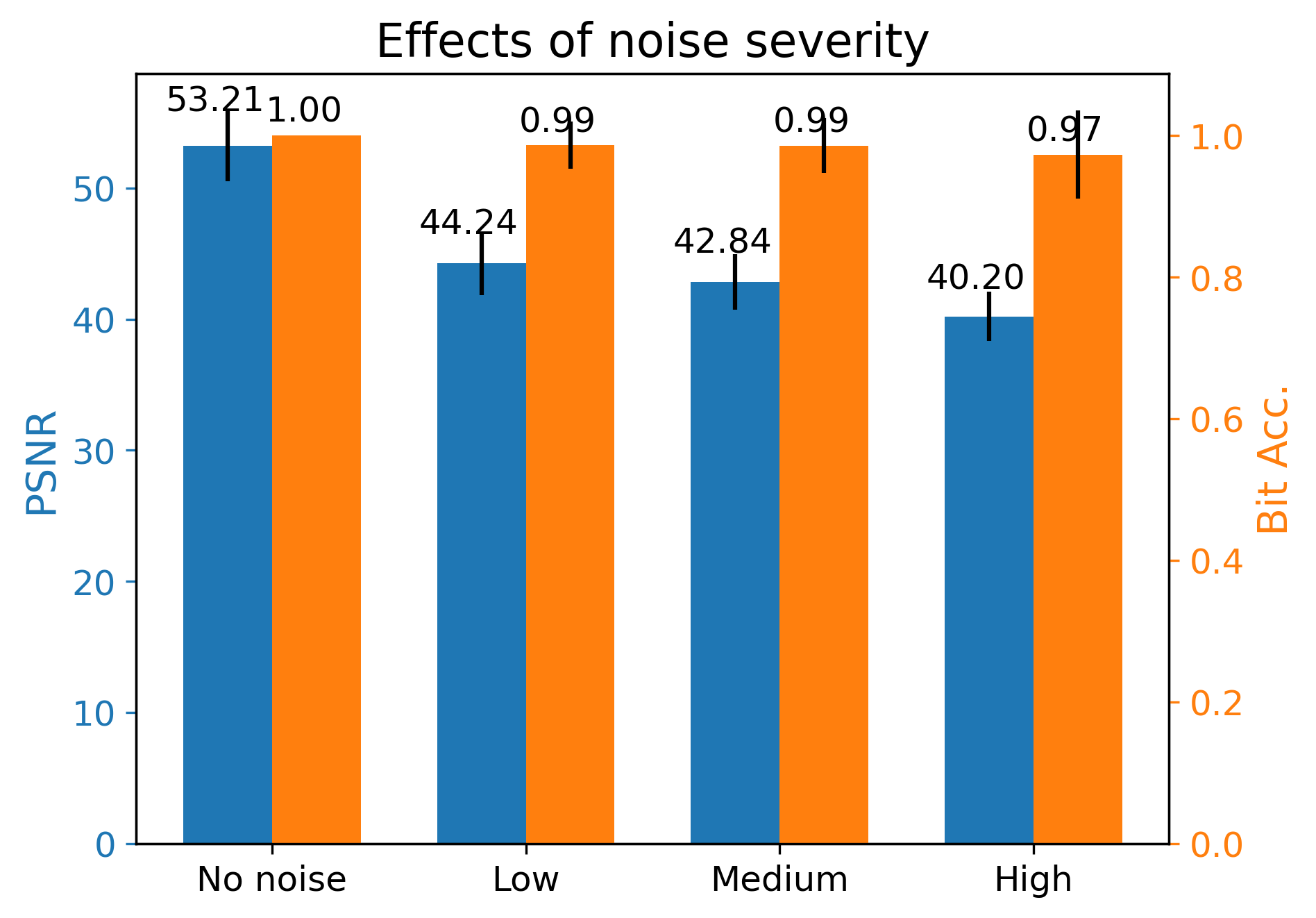} &
    \includegraphics[width=0.3\linewidth,trim=0cm 0cm 0cm 0cm,clip]{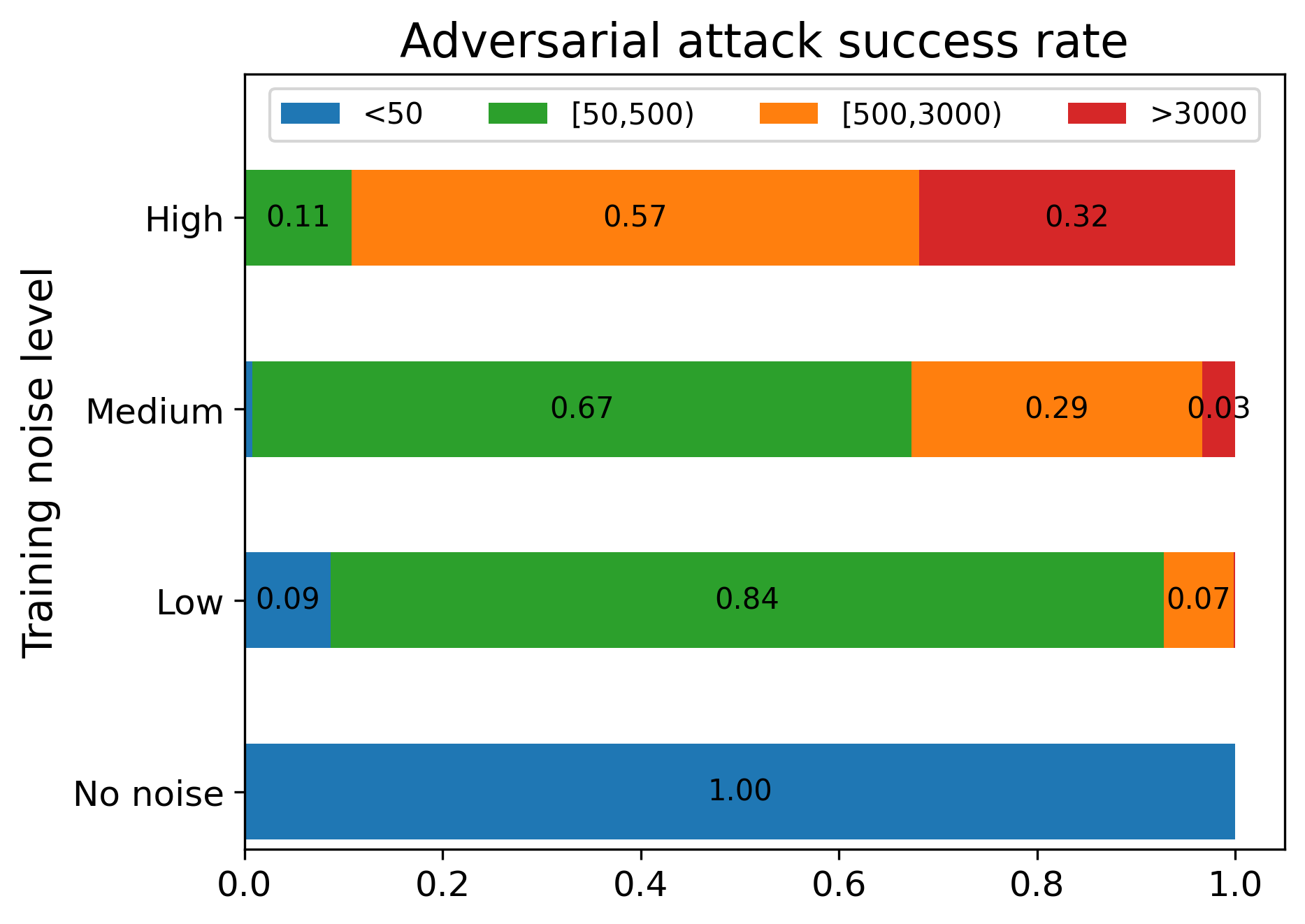} \\
    (a) Bit length & (b) Noise severity & (c) Adversarial attack 
    \end{tabular}
    \caption{Impact of bit length, noise severity and adversarial attack on TrustMark performance on DIV2K. Larger watermark reduces both PSNR and bit accuracy (a). Training with higher noise severity affects PSNR more than bit accuracy (b); and improves robustness against adversarial attack (c).}
    \squeezeup
    \label{fig:robust}
\end{figure*}


Since the watermark signal is small as opposed to the image content, it is important to prioritize the watermark extractor $X$ at the early training phase. We set the trade-off parameter $\alpha$ low initially ($\alpha=0.05$) and disable noise simulation and GAN loss as well as repeating the first image batch while varying random watermarks until $X$'s detection accuracy exceeds a certain threshold. We then unlock subsequent TrustMark features before increasing $\alpha$ to the intended value $\alpha_{\mathrm{max}}$ (more details in Sup.Mat.).  

\noindent \textbf{Metrics}. We use standard PSNR and Structure Similarity Index Measure (SSIM) metrics for evaluating imperceptibility our watermarking/removal algorithms; and bit accuracy for watermark recovery (50\% for random guess). Unless otherwise stated, we apply \cref{alg:arbitrary} for the proposed methods and all baselines and compute these metrics at the original image resolution.

\noindent \textbf{Baselines}. We compare TrustMark with recent watermark and steganography baselines including RoSteALS \cite{bui2023rosteals}, RivaGAN \cite{zhang2019robust}, SSL \cite{fernandez2022watermarking}, StegaStamp \cite{tancik2020stegastamp} and dwtDctSvd \cite{navas2008dwt}. For fair comparison, we retrain the baselines using noise simulation settings as TrustMark (\cref{sec:noise}) {\em if} it helps to improve performance. We also report TrustMark performance on their reported settings. At inference, Resolution Scaling is applied to all methods except RivaGAN and dwtDctSvd which support native resolution.

\begin{figure*}[]
    \centering
    \includegraphics[width=1.0\linewidth,trim=0cm 0cm 0cm 0cm,clip]{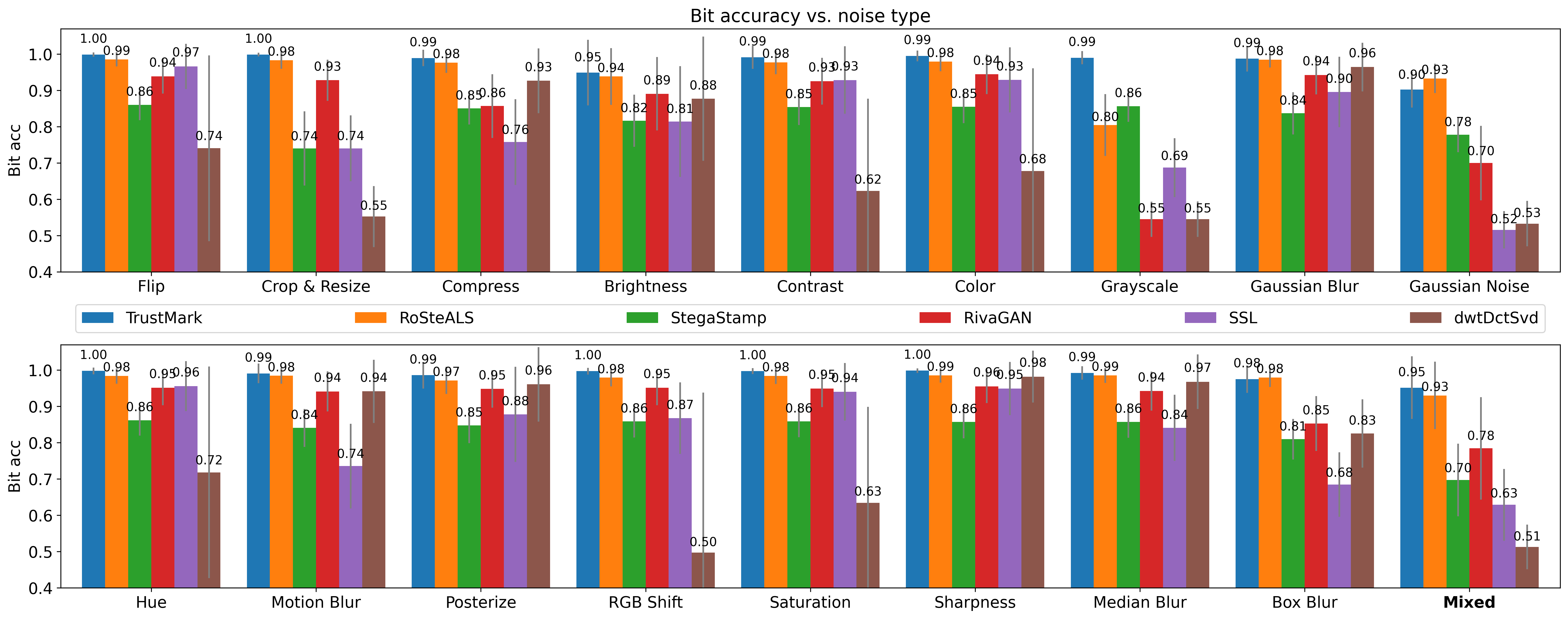}
    \squeezeup
    \caption{TrustMark-Q ($\alpha_{\mathrm{max}}=27.5$) robustness to individual noise perturbations, evaluated on the DIV2K dataset.}
    \squeezeup
    \label{fig:ind_noise}
\end{figure*}

\subsection{Watermark embedding}
\label{exp:emb}
\begin{table}[t!]
\centering
\definecolor{Gray}{gray}{0.9}
\small
\begin{tabular}{lccc}
\toprule
Method & PSNR$\uparrow$ & SSIM$\uparrow$ & Bit Acc. ($\sim 0.5$)\\
\midrule
TrustMark-RM & 48.48 & 0.997 & 0.553\\
I-FGSM \cite{ifgsm} & 23.48 & 0.613 & 0.629\\
\bottomrule
\end{tabular}
\caption{Watermark removal for TrustMark-B, on DIV2K.}
\label{tab:wr}
\squeezeup
\end{table}

\cref{tab:main} shows performance of TrustMark and other baselines on the three benchmarks. We report two TrustMark versions depending on the choice of the trade-off parameter: TrustMark-B ($\alpha_{\mathrm{max}}=20$) with balanced image quality and watermark recovery, and TrustMark-Q ($\alpha_{\mathrm{max}}=27.5$) with prioritized image quality. Overall, TrustMark-Q outperforms all baselines at every metrics on all benchmarks except PSNR on DIV2K comparable with SSL. TrustMark-B lags behind TrustMark-Q $\sim$2dB on PSNR but leads by 1-2\% for bit accuracy on noisy watermarked images. TrustMark, RoSteALS and dwtDctSvd achieve near perfect bit accuracy on clean images. When random noises are added, dwtDctSvd and StegaStamp see the greatest drops in performance. Benchmark-wise, DIV2K proves to be the most challenging for watermarking, while the narrow-domain MetFace yields the highest performance for all methods. Interestingly, SSL bit accuracy (clean) is perfect on MetFace but poor on other two benchmarks, indicating it is not robust to resize that does not keep the image aspect ratio (all MetFace images are square).

\begin{figure*}[ht]
    \begin{minipage}[b]{.55\linewidth}
    \centering
    \small
    \begin{tabular}{lll|cc|cc}
\toprule
GAN  & FFL & $\textbf{E}_{\mathrm{post}}$ & PSNR           & SSIM            & Acc. (clean)    & Acc. (noised)   \\ \midrule
\checkmark    & \checkmark   & \checkmark                   & \textbf{40.203} & \textbf{0.987}   & 0.999 & 0.973\\
     & \checkmark   & \checkmark                   & 39.603& 0.985 & 0.999 & 0.975 \\
\checkmark    &     & \checkmark                   & 39.526 & 0.985 & 0.999 & 0.974 \\
\checkmark    & \checkmark   &                     & 38.366 & 0.983 & 0.998 & 0.970 \\
     &     & \checkmark                   & 38.916  & 0.983 & 1.00            & 0.979 \\
\checkmark    &     &                     & 37.034 & 0.978 & 0.989 & 0.963\\
     & \checkmark   &                     & 38.084 & 0.981 & 0.999& 0.974\\
     &     &                     & 36.300 & 0.976 & 0.991 & 0.959\\ \hline
\multicolumn{3}{l|}{RegNet \cite{regnet}}      & 39.259 & 0.984& \textbf{1.00}            & \textbf{0.986} \\
\multicolumn{3}{l|}{ResNext \cite{resnext}}     & 39.461 & 0.985 & 1.00            & 0.976 \\
\multicolumn{3}{l|}{ResNet18 \cite{resnet}}    & 38.856  & 0.983 & 0.998 & 0.972 \\
\multicolumn{3}{l|}{DenseNet121 \cite{densenet}} & 39.569  & 0.985 & 1.00            & 0.982\\
\bottomrule
    \end{tabular}
    \captionof{table}{Ablation studies on different mutations of TrustMark ($\alpha_{\mathrm{max}}=20$) architecture, evaluated on DIV2K.}
    \label{tab:abl}
  \end{minipage}\hfill
  \begin{minipage}[b]{.4\linewidth}
    \centering
    \small
    \includegraphics[width=1.0\linewidth,trim=0cm 0cm 0cm 0cm,clip]{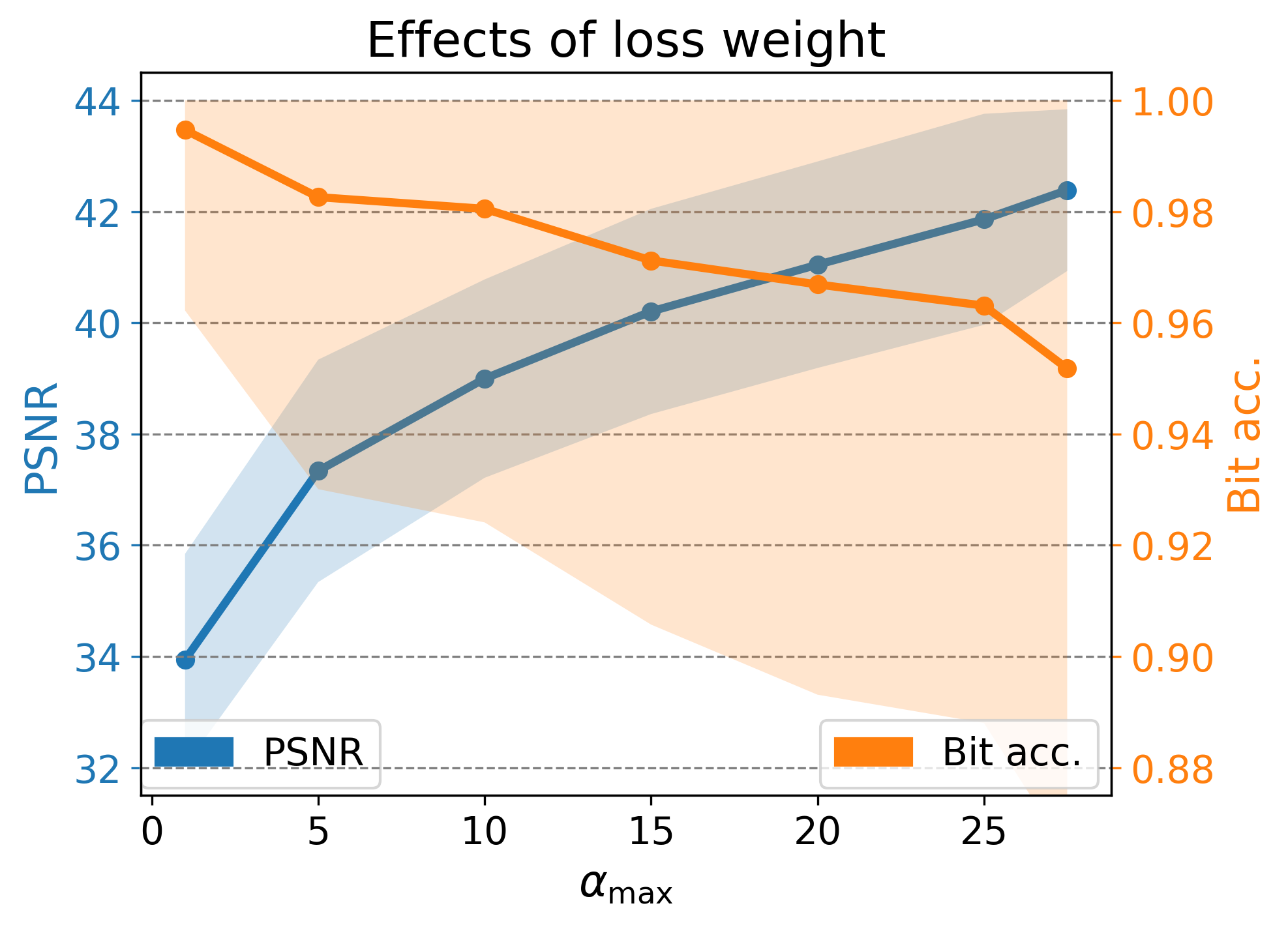}
    \caption{Trade-off between imperceptibility and watermark recovery, evaluated on DIV2K.}
    \label{fig:tradeoff}
  \end{minipage}
  \squeezeup
\end{figure*}

\cref{fig:tradeoff} depicts TrustMark performance at various $\alpha_{\mathrm{max}}$ values, confirming the trade-off between imperceptibility and watermark recovery. We note the optimal range of $\alpha_{\mathrm{max}}$ is (0,30) for stable training. As $\alpha_{\mathrm{max}}$ increases to 30, the PSNR improves by more than 8dB while bit accuracy drops by less than 5\%. \cref{fig:eg} shows watermarking results for several cover images. TrustMark's artifacts are more uniform in color (RGB residual is a gray image), more invariant to semantic objects (it is harder to recognize image objects from the residual) and is overall less visible than other methods. 

\begin{figure*}[]
    \centering
    \includegraphics[width=1.0\linewidth,trim=0cm 0cm 0cm 0cm,clip]{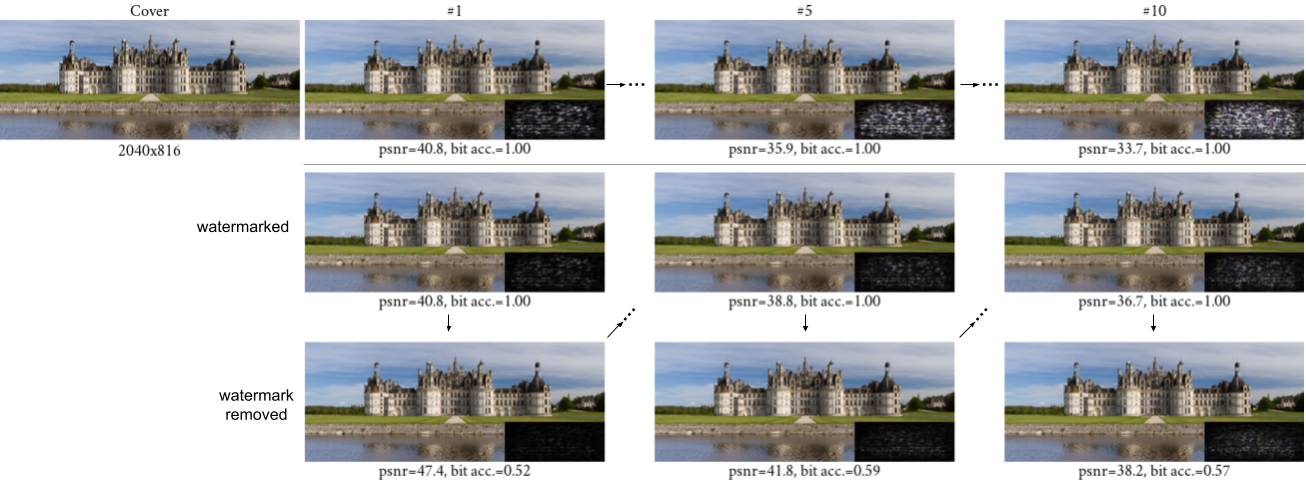}
    \squeezeup
    \caption{Re-watermarking example. First row - left to right: cover image and encoded images after re-watermarking 1,5 and 10 times with random watermarks. Second and third rows: TrustMark-RM is applied to remove artifact before re-watermarking. Inset: $20 \times$ residual.}
    \squeezeup
    \label{fig:wr_eg}
\end{figure*}
We also train and evaluate TrustMark using the same settings as the recent work of RoSteALS \cite{bui2023rosteals}. Specifically, the noise model \textbf{N}(.) is replaced by ImageNet-C perturbations \cite{imagenetc} and resolution scaling (\cref{alg:arbitrary}) is turned off, meaning all performance metrics are computed at the model-designed resolution. We note that ImageNet-C noises are often more severe than \textbf{N}(.) but lack geometric transformations. \cref{tab:imgc} shows that TrustMark (optimal $\alpha_{\mathrm{max}}=15$) achieves better watermark recovery than RoSteALS and much better PSNR score (RoSteALS PSNR is capped by its frozen autoencoder at 33.9dB). TrustMark's PSNR is comparable with RivaGAN and dwtDctSvd but its watermark recovery performance is significantly higher. Meanwhile, SSL has the highest PSNR with bit accuracy among the lowest, indicating vulnerability to noise.   

\subsection{Watermark size and robustness}
\label{sec:robust}

\cref{fig:robust} (a) shows TrustMark-B performance for the bit length range of [32,200]. Overall, it is more challenging to embed and decode large watermarks, as PSNR and bit accuracy both drop by 7.5dB and 11\% when bit length increases 6 folds from 32 to 200, respectively.

We assess TrustMark robustness on several facets -- against various noise sources, noise severity levels and adversarial attack. \cref{fig:ind_noise} shows bit accuracy performance of TrustMark ($\alpha_{\mathrm{max}}=27.5$) and other baselines against every individual noise sources in \cref{sec:noise}. TrustMark outperforms the closest baseline RoSteALS on all sources except Gaussian noise and box blur. Other methods are robust against certain noises but weak against others \eg dwtDctSvd performance is close to random chance for RGB shift. For noise severity, we train and test TrustMark-B on 3 additional variants of noise settings: no noise simulation, low-level noise and medium-level noise. 
\cref{fig:robust}(b) demonstrates that increasing noise severity affects PSNR the most while bit accuracy stays roughly the same. Specifically, PSNR sets at 53.2dB without noise simulation then drops to 40.2dB for high severity noise, but bit accuracy reduces by only 3\%. Using high severity noise simulation during training also makes TrustMark more robust to adversarial attack, as shown in \cref{fig:robust}(c). Here, we perform I-FGSM attack \cite{ifgsm} by adding subtle noise with maximum strength $\epsilon=8/255$ into the watermarked image to fool the watermark extraction model. The adversarial noise is initially set to 0 then is adjusted at each attack iteration until the bit accuracy of the target image is brought down below $0.5+\epsilon/2$. The number of I-FGSM iterations reflects the robustness of the watermark model against adversarial attack. Per \cref{fig:robust}(c), when TrustMark is trained without noise simulation, it takes less than 50 iterations for a successful attack on any watermarked image. In contrast, when trained with high level noises, 32\% of the watermarked images require more than 3000 attack iterations. 

\subsection{Watermark removal and re-watermarking}
\label{sec:exp_wr}

Without previous work about watermark removal, we compare our method with adversarial attack I-FGSM \cite{ifgsm}. \cref{tab:wr} shows that TrustMark-RM not only brings down bit accuracy to 55\% but also improves PSNR to 48.5dB. In contrast, I-FGSM  worsens the image quality as its sole aim is to `break' the decoder module. Note that both methods do not achieve random chance bit accuracy because the resolution scaling post-process weakens its efficacy. 

\cref{fig:rewm} demonstrates TrustMark-RM efficacy for re-watermarking. We make 2 observations: (i) bit accuracy is not affected regardless whether a watermark remover is employed (\cref{fig:rewm} bottom); and (ii) TrustMark-RM preserves image quality better than not using it (\cref{fig:rewm} top). However, the denoising effect of TrustMark-RM is weaken after each re-watermarking time, because the unwanted noise generated by TrustMark-RM gets accumulated over times. A re-watermarking example is illustrated in \cref{fig:wr_eg}.

\begin{figure}[t]
    \centering
    \includegraphics[width=1.0\linewidth,trim=0cm 0cm 0cm 0cm,clip,height=5.2cm]{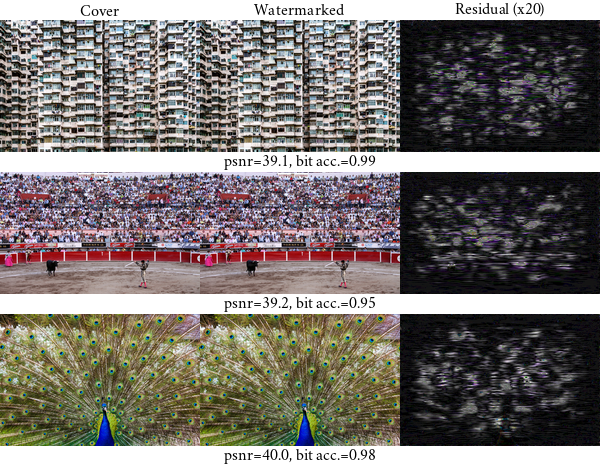}
    \squeezeup
    \caption{Limitations. Cluttered images are harder to watermark, both in the embedding and watermark extraction tasks resulting in slightly lower accuracy and PSNR.}
    \squeezeup
    \label{fig:limit}
\end{figure}

\subsection{Ablation study}
\label{sec:ablation}

\cref{tab:abl} shows the contribution of each design components. Our training strategy ensures watermark recovery is always prioritized in the first training phase, enabling bit accuracy performance to be maintained through various ablations. The architecture mutations influence PSNR mostly. When GAN, $\textbf{E}_{\mathrm{post}}$ and FFL loss are all disabled, TrustMark PSNR is equivalent to StegaStamp. Adding GAN, $\textbf{E}_{\mathrm{post}}$ and FFL separately improves the score by 0.7dB, 1.6dB and 1.7dB respectively. $\textbf{E}_{\mathrm{post}}$ combined with either GAN or FFL boosts PSNR by 3dB and all three components make up 4dB in PSNR and 1.4\% in bit accuracy in total.

We also experiment with different backbones for our watermark decoder \textbf{X}. TrustMark training converges for ResNet family, including ResNet18 \cite{resnet}, DenseNet121 \cite{densenet}, RegNet \cite{regnet} and ResNext \cite{resnext}, but is not successful for VGG \cite{vgg}, GoogleNet \cite{googlenet}, ConvNext \cite{convnext} and EfficientNet \cite{efficientnet}. We observe all successful backbones have either a residual layer or a skip layer - both allow signals from bottom layers to flow directly to the top via a sum (residual) or concatenation (skip) operation. We attribute this unique requirement of TrustMark to the combination of watermarking nature (subtle watermark signal), the complexity of our multi-noise simulation scheme and the accuracy-thresholded multi-stage training procedure (see Sup.Mat.).

Finally we evaluated high and arbitrary resolution watermarking by ablating the DIV2K dataset to 20\% to 100\% of original (2K) resolution.  We observe after encoding that PSNR varies only by $\pm 0.02$dB and after decoding that bit accuracy varies only by $\pm 10^{-4}$ on average across all resolutions.  We conclude TrustMark shows near-equivalent performance across arbitrary resolutions due to our Resolution Scaling technique (\cref{sec:arbitrary_res}), which we contrast against simple bi-linear/cubic interpolation in the Sup. Mat.

\subsection{Limitations}
\noindent We investigate TrustMark weaknesses by examining the encoded images with the lowest PSNR scores. \cref{fig:limit} shows several examples; all are highly cluttered. We hypothesize that the subtle watermarks interfere with the high-frequency content components which we enhanced preservation of and which dominate in cluttered images, and consequently slightly reduce the PSNR score and also the TrustMark decoder's capability in recognizing it. Given the small number of wrongly detected bits, this is of no practical significance as a bit errors can be alleviated using a forward-error correction method such as BCH \cite{bch} at the cost of shortening effective payload in the watermark. We do not apply such correction in any of our reported results.

\section{Conclusion}
\label{sec:conclusion}
\noindent We propose TrustMark and TrustMark-RM for watermarking and watermark removal. TrustMark integrates novel designs in the post-processing layers together with a focal frequency loss and rigorous noise simulation for robustness. With TrustMark-RM, we show for the first time that imperceptible watermark can be treated as noise, therefore removable using a denoising network. We propose an effective resolution scaling algorithm to extend TrustMark and TrustMark-RM for images with arbitrary resolution. We show TrustMark encoded images are imperceptible (PSNR$>40$dB) while being state-of-art robust to various noise sources and can be restored with high fidelity (PSNR$>48$dB). Future applications of this work  include creative workflows where identifiers may be imperctibly embedded to track the provenance of image assets in conjuction with emerging standards \cite{c2pa}.

\noindent\textbf{Acknowledgment} This work was supported in part by DECaDE under EPSRC grant EP/T022485/1.
{\small
\bibliographystyle{ieee_fullname}
\bibliography{egbib}
}

\clearpage
\begin{appendices}

\section{More training details}
\subsection{Training stages}
\begin{figure}[t!]
    \centering
    \includegraphics[width=1.0\linewidth]{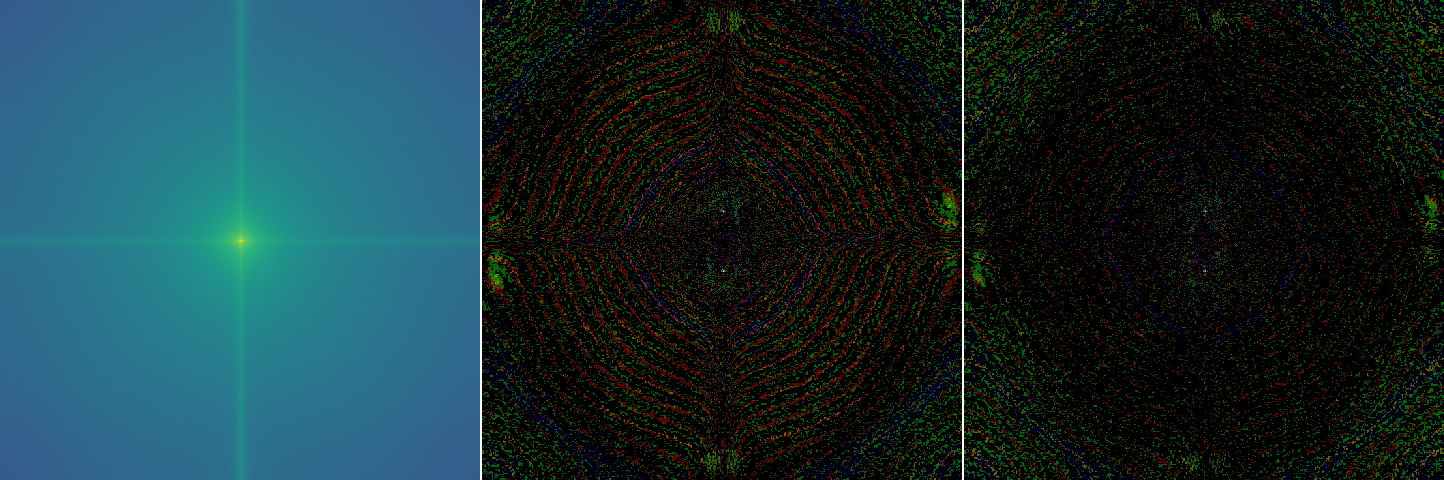}
    \squeezeup
    \caption{Effects of FFL in frequency domain: (left) average frequency spectrum of cover images in the MetFace benchmark, (middle) residual spectrum without FFL and (right) with FFL. The residual images are amplified 100x for visualization purpose.}
    \squeezeup
    \label{fig:freq_ffl}
\end{figure}
Our training of TrustMark consists of 4 stages in an end-to-end manner to encourage convergence (Sec. 4.1 Training details, main paper). \cref{tab:phase} details the components to be activated in each stage and the triggering thresholds. At stage 0, we supply a fixed training batch of cover images to the network at every iteration and only randomize the watermarks. The noise model \textbf{N} and GAN loss $\mathcal{L}_{\mathrm{GAN}+\mathrm{GP}}$ are turned off; the trade-off parameter $\alpha$ is set at a low value of 0.05 to prioritize training the watermark extractor \textbf{X}. When the training bit accuracy reaches 90\%, random image batches of the entire training dataset are fed into the network (stage 1). Noise simulation is activated when a training batch achieves 95\% accuracy (stage 2). Finally, at bit accuracy of 98\%, the GAN loss is turned on and $\alpha$ linearly increases to $\alpha_{\mathrm{max}}$ (20 for TrustMark-B and 27.5 for TrustMark-Q) for the next 10,000 iterations. We find these 4-stage settings aid convergence and are particularly useful when  high-severity noise simulation, large $\alpha_{\mathrm{max}}$ or watermark size are required. A typical training of TrustMark-B takes 500, 600 and 800 iterations to reach stage 1, 2 and 3 respectively.

\begin{figure}[]
    \centering
    \includegraphics[width=0.9\linewidth]{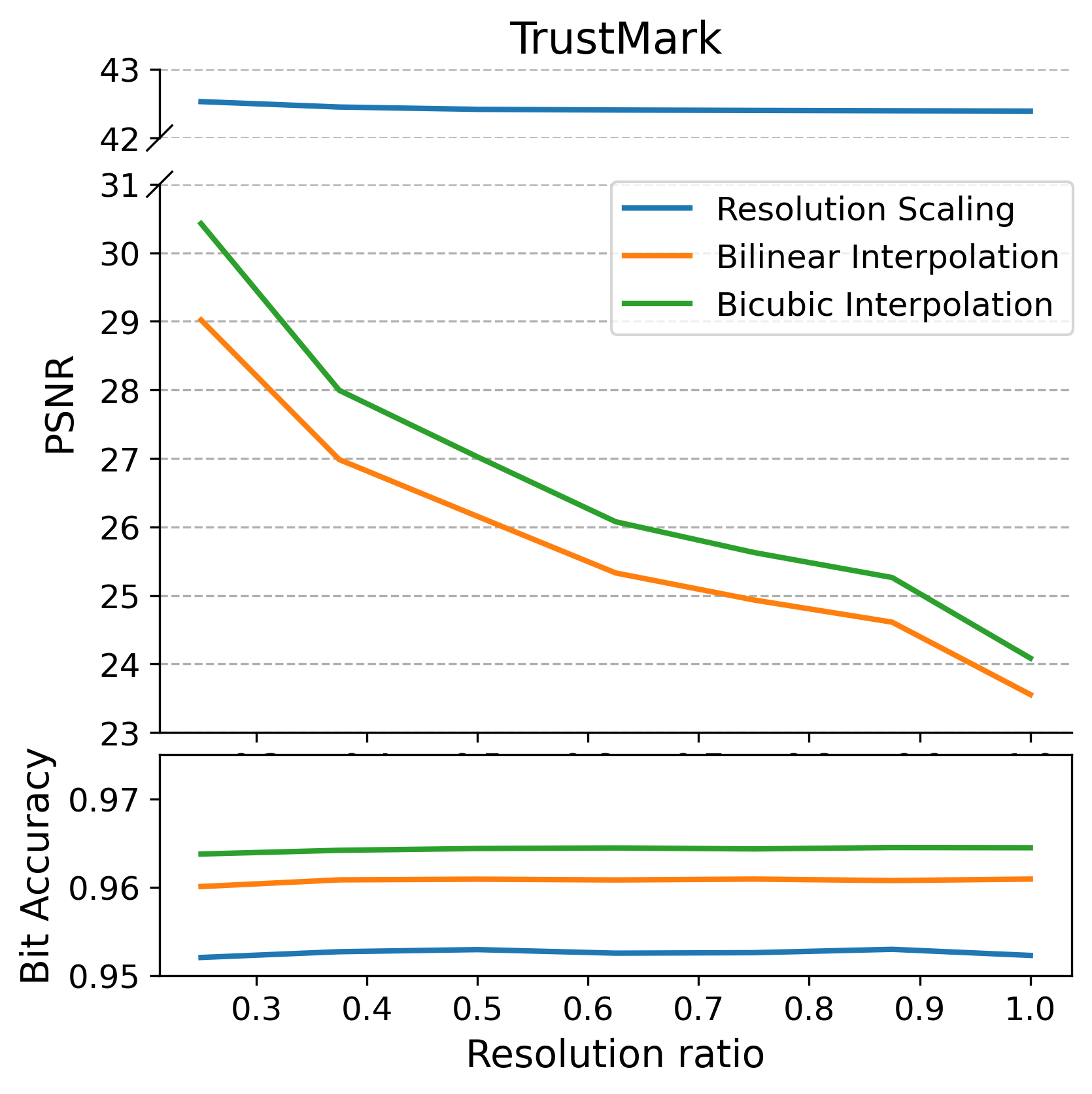}
    \squeezeup
    \caption{Resolution Scaling performance versus Bilinear and Bicubic interpolations for the watermarking task, evaluated with TrustMark ($\alpha_{\mathrm{max}}=27.5$) on DIV2K at different resolutions. }
    \label{fig:multires}
\end{figure}

\begin{figure}[t!]
    \centering
    \includegraphics[width=0.9\linewidth]{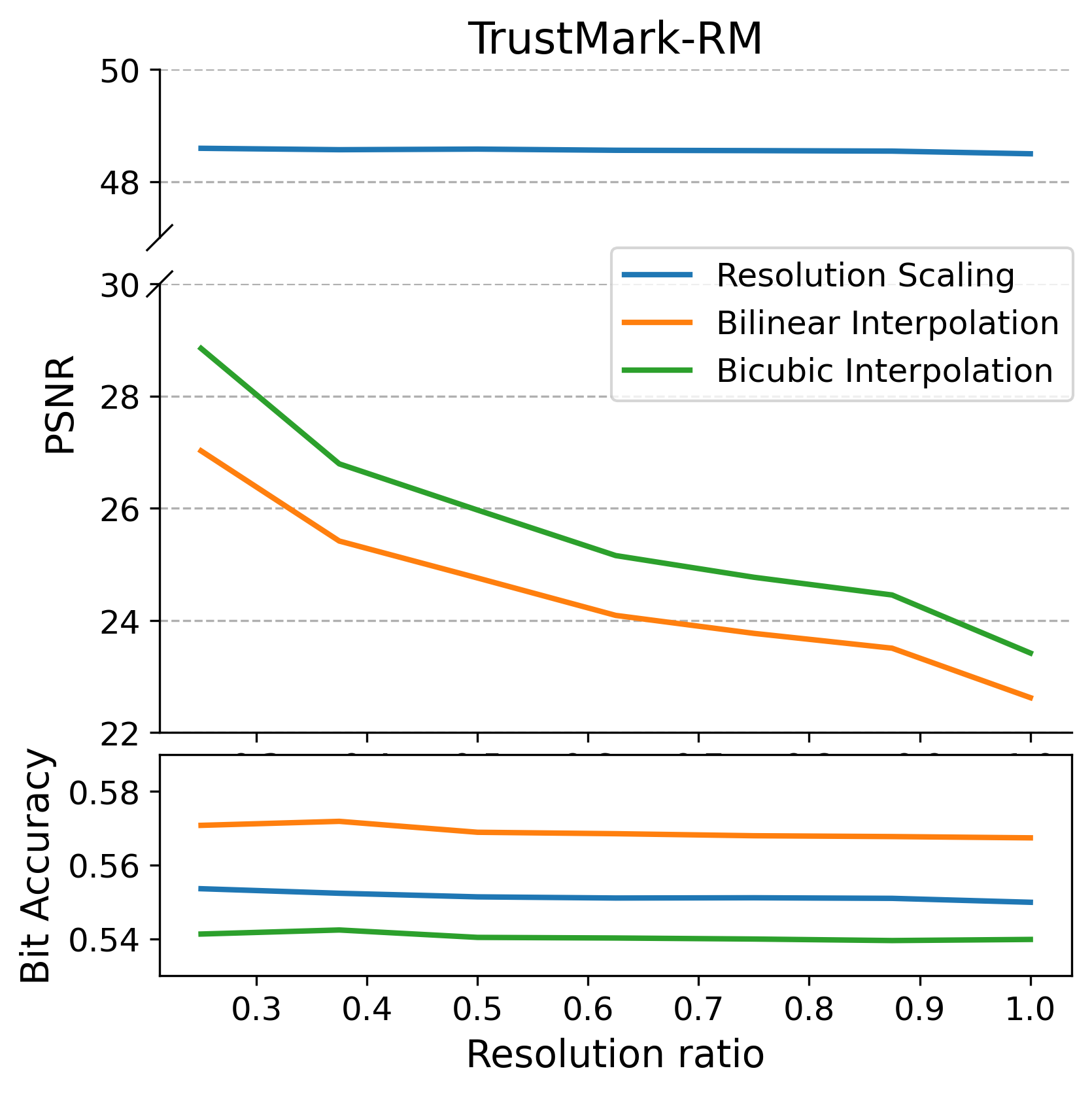}
    \squeezeup
    \caption{Resolution Scaling performance versus Bilinear and Bicubic interpolation for the watermarking removal task, evaluated with TrustMark-RM on DIV2K.}
    \label{fig:multires_rm}
\end{figure}

\begin{table*}[]
\centering
\begin{tabular}{l|ccccc|c}
\toprule
\multirow{2}{*}{Stage} & \multicolumn{5}{c|}{Settings} & Triggered \\
 & Fixed batch & Rnd. batch & Noise simulation & $\alpha$ ramping & GAN & bit acc. \\
 \midrule
0 & \checkmark &  &  &  &  & - \\
1 &  & \checkmark &  &  &  & 0.90 \\
2 &  & \checkmark & \checkmark &  &  & 0.95 \\
3 &  & \checkmark & \checkmark & \checkmark & \checkmark & 0.98\\
\bottomrule
\end{tabular}
\caption{Four stages of TrustMark in an end-to-end training pipeline. The next stage is automatically triggered once a training batch bit accuracy exceed a corresponding threshold.}
\label{tab:phase}
\end{table*}

\subsection{Noise severity}
\cref{tab:noise} shows details of our noise settings (c.f. Fig. 4 (b) and Sec. 4.3, main paper). We employ the Kornia library \cite{kornia} for implementation of the differentiable transformations, except for Jpeg Compression in which the implementation algorithm comes from \cite{shin2017jpeg}. The geometrical transformations (not shown in \cref{tab:noise}) are kept fixed in all settings, specifically random flip with probability of 50\%, random resize with scale range 0.8--1.0 and aspect ratio range 3/4--4/3 and crop size of 244 out of 256.  Examples of the noises are shown in \cref{fig:noise_eg}.
\begin{figure*}[t!]
    \centering
    \includegraphics[width=1.0\linewidth,trim=0cm 0cm 0cm 0cm,clip]{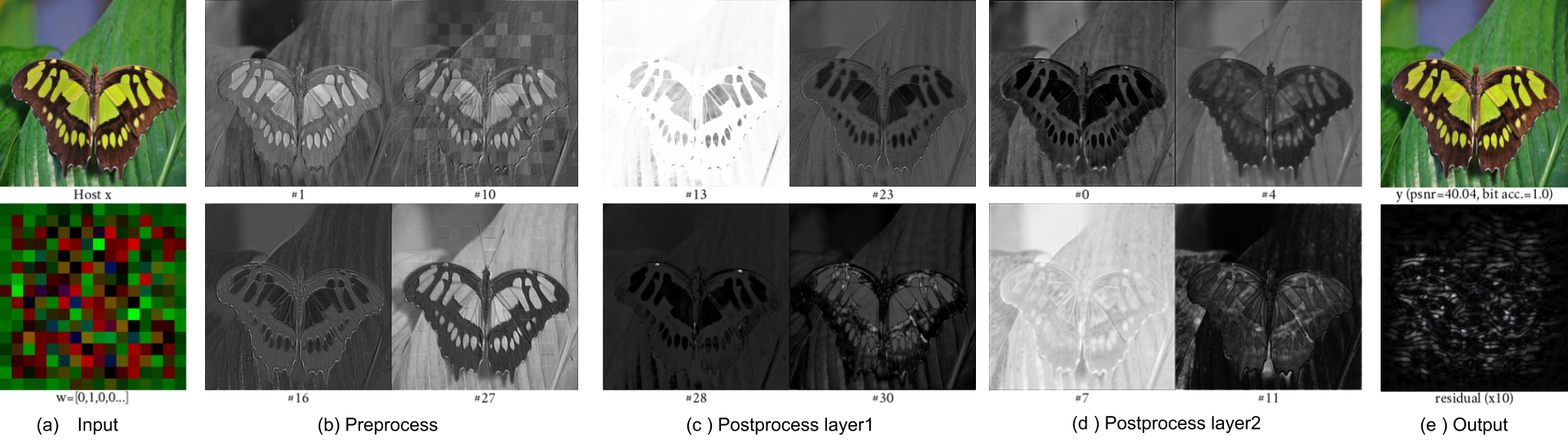}
    \squeezeup
    \caption{TrustMark pre- and post-process activation maps, best with zoom-in.}
    \squeezeup
    \label{fig:pre}
\end{figure*}

\begin{figure*}[t!]
    \centering
    \includegraphics[width=1.0\linewidth,trim=0cm 0cm 0cm 0cm,clip]{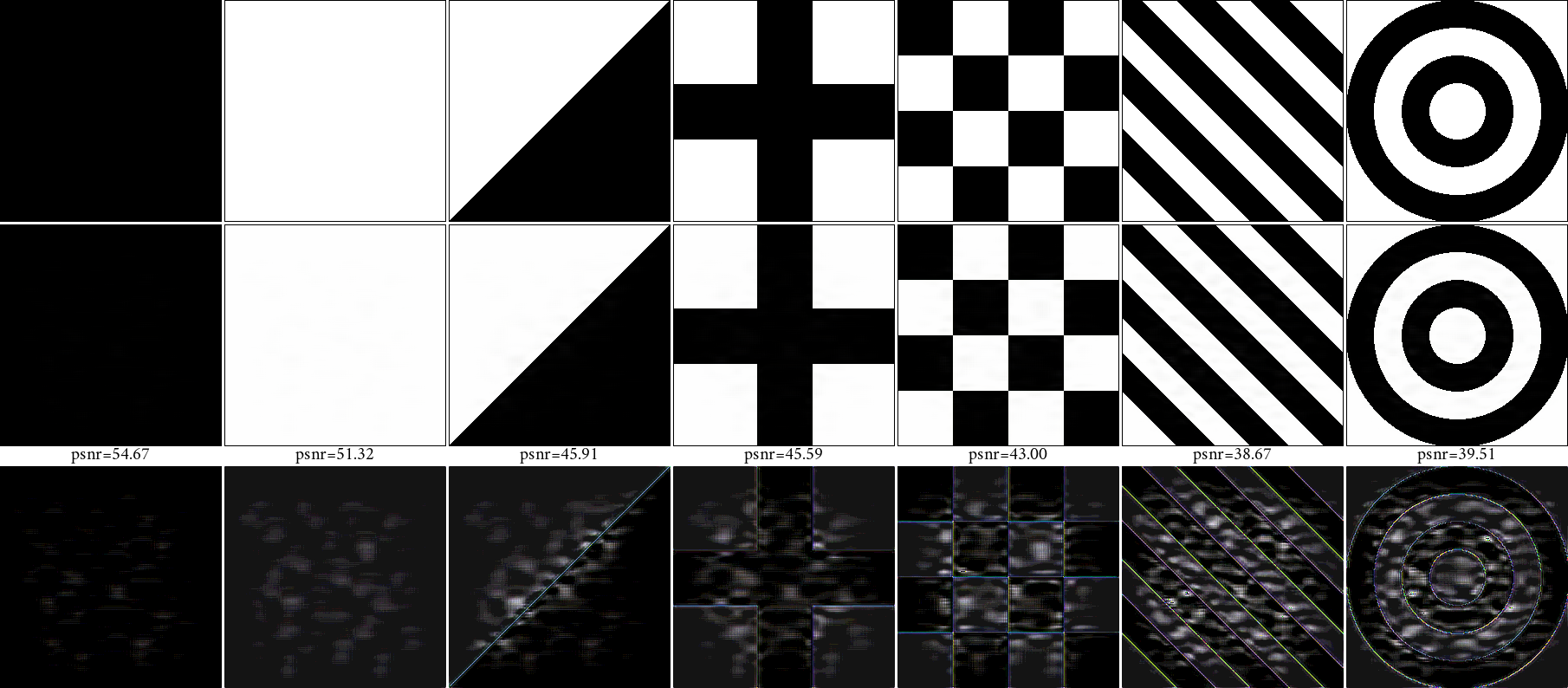}
    \squeezeupsmall
    \caption{Watermarking on a set of control images. All achieves 100\% bit accuracy. First row: host images , middle row: encoded images with a same random watermark; bottom row: residual images scaled up 20x.}
    \label{fig:eg_control}
\end{figure*}

\begin{table*}[]
\centering
\begin{tabular}{l|ccccccc}
\toprule
 & Jpeg Compress. & Brightness & Contrast & Color Jiggle & Grayscale & Gaussian Blur & Gaussian Noise \\
 & (min q) & (bri.) & (con.) & (bri., con., sat., hue) & (p) & (k, $\sigma$) & (std.) \\
 \midrule
Low & 70 & 0.9--1.1 & 0.9--1.1 & 0.05, 0.05, 0.05, 0.01 & 0.5 & 3, 0.1--1.0 & 0.02 \\
Medium & 50 & 0.75--1.25 & 0.75--1.25 & 0.1, 0.1, 0.1, 0.02 & 0.5 & 5, 0.1--1.5 & 0.04 \\
High & 40 & 0.5--1.5 & 0.5--1.5 & 0.1, 0.1, 0.1, 0.05 & 0.5 & 7, 0.1--2.0 & 0.08\\
\bottomrule
\end{tabular}
\\
\begin{tabular}{l|cccccccc}
\toprule
 & Hue & Posterize & RGB Shift & Saturation & Sharpness & Median blur & \multicolumn{1}{l}{Box blur} & Motion blur \\
 & (hue) & (bits) & (shift limit) & (sat.) & (sha.) & (k) & (k) & (k,$\beta$, t) \\
 \midrule
Low & 0.01 & 5 & 0.02 & 0.9-1.1 & 0.5 & 3 & 3 & 3-5,-25--25,-0.25--0.25 \\
Medium & 0.02 & 4 & 0.05 & 0.75-1.25 & 1.0 & 3 & 5 & 3-7,-45--45,-0.5--0.5 \\
High & 0.05 & 3 & 0.1 & 0.5-1.5 & 2.5 & 3 & 7 & 3-9,-90--90,-1.0--1.0\\
\bottomrule
\end{tabular}
\caption{Noise simulation settings at low, medium and high levels. `a--b' refers to the value range between a and b that a parameter is randomly drawn from. For Jpeg compression, q is the compression factor (or image quality). For blurring operations, k is the kernel size. Additionally, for motion blur, $\beta$ and t are the angle and direction of the simulated motion. \em{bri., con., sat.} and \em{sha.} are the brightness, contrast, saturation and sharpness factors in Kornia \cite{kornia}.} 
\label{tab:noise}
\end{table*}

\section{Resolution Scaling robustness}

\begin{figure*}[t]
    \centering
    \includegraphics[width=0.9\linewidth]{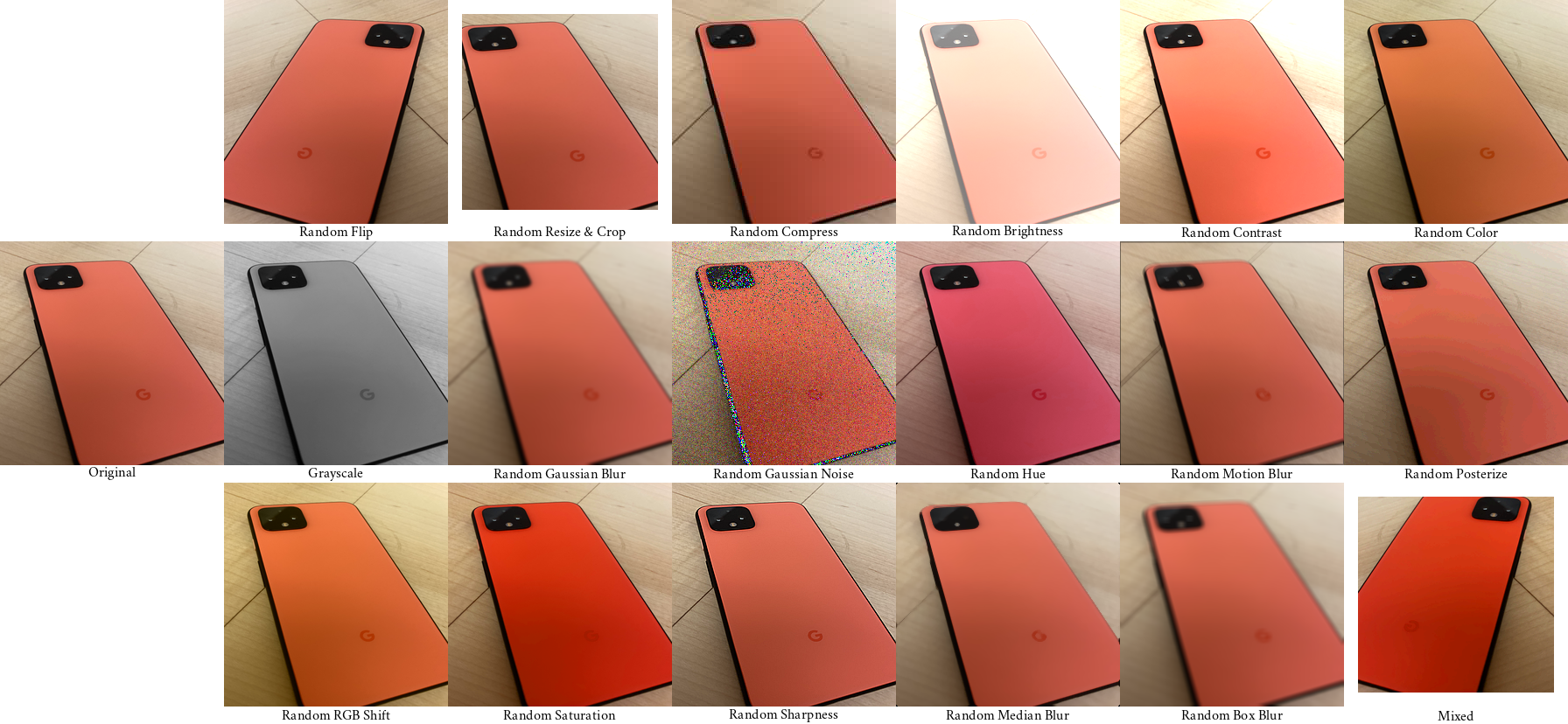}
    \caption{Examples of different noise sources applied during TrustMark training. }
    \label{fig:noise_eg}
\end{figure*}

We examine the robustness of Resolution Scaling against resolution change in term of preserving image quality and watermark recovery. We first create multiple versions of the DIV2K benchmark by downsampling every image with a factor $0< \rho <1$, then evaluate our watermark embedding (TrustMark, \cref{fig:multires}) and removal (TrustMark-RM, \cref{fig:multires_rm}) on these newly created benchmarks. We also compare with Bilinear and Bicubic interpolation methods \ie output images at 256$\times$256 resolution are scaled to the target resolution via Bilinear or Bicubic samplings instead of Resolution Scaling.  \cref{fig:multires} and \cref{fig:multires_rm} show that all methods drop performance in PSNR when the target resolution increases beyond 256$\times$256 towards the original image resolution, however the drop for Resolution Scaling is negligible (0.1dB for both TrustMark and TrustMark-RM) as compared with Bilinear and Bicubic interpolations. On the other hand, the bit accuracy for Resolution Scaling is lower than Bicubic by 1\% for TrustMark (higher is better) and TrustMark-RM (lower is better).

\noindent \textbf{Note on PSNR's resolution dependence}. We conducted an experiment to assess the dependence of the PSNR metric to image resolution (\cref{fig:psnr_dep}). Here, we compute PSNR on a pair of 2 totally different images ($\mathrm{x}_1, \mathrm{x}_2$) and a pair of cover and watermarked images ($\mathrm{x}_1, \hat{\mathrm{x}}_1$). All images $\mathrm{x}_1$, $\mathrm{x}_2$ and $\hat{\mathrm{x}}_1$ have the same resolution of 512$\times$512 and the watermarking algorithm is dwtDctSvd \cite{navas2008dwt}. PSNR is computed at the original resolution, and also at various downsampling and upsampling resolutions of the 2 pairs. While the PSNR scores for ($\mathrm{x}_1, \mathrm{x}_2$) vary slightly with 0.35dB maximum difference, the change in PSNR for ($\mathrm{x}_1, \hat{\mathrm{x}}_1$) is significantly higher (4.7dB). This behavior does not depend on the interpolation (Bilinear, Bicubic) nor the watermarking methods. This indicates that PSNR is more sensitive to similar image pairs (as is the case of watermarking) when evaluated at different resolutions. It is because the subtle watermark degrades after up/down-sampling, increasing PSNR in overall. In the broader context, if two watermark models work on two different image resolutions and are evaluated at its model-designed resolution, the resulting PSNR scores will not be directly comparable. This highlights the contribution of our Resolution Scaling algorithm in extending fixed-resolution watermarking methods to work on original resolution, therefore can be directly compared with each other and with other arbitrary-resolution methods.  
\section{Trade-off control at inference time}
\begin{figure}[]
    \centering
    \includegraphics[width=0.9\linewidth]{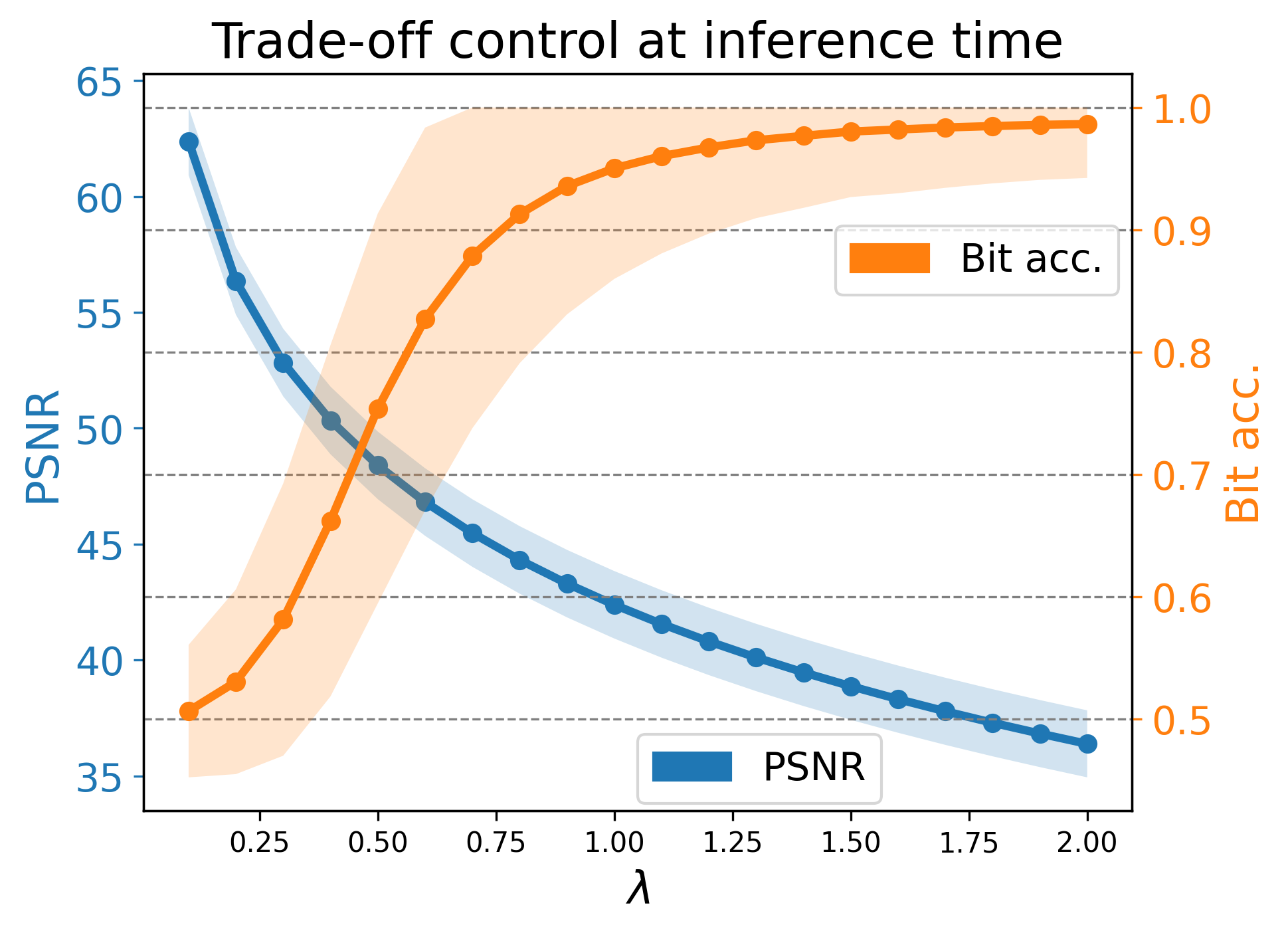}
    \squeezeup
    \caption{Control of the imperceptibility and watermark recovery trade-off can be achieved at inference time via the parameter $\lambda$ in \cref{eq:rescale}, tested on DIV2K.}
    \label{fig:rescale}
\end{figure}
We demonstrate in Sec. 4.2 and Fig. 6 of the main paper that the trade-off between imperceptibility and watermark recovery can be controlled via the loss weight parameter $\alpha_{\mathrm{max}}$, which must be decided at training time. Here, we show that the trade-off can also be controlled at inference time by inserting a control parameter $\lambda>0$ to L\#9 of Alg. 1;
\begin{equation}
    \mathbf{y} \leftarrow \mathrm{clamp}(\mathbf{x} + \lambda*\mathbf{r}, -1, 1)
    \label{eq:rescale}
\end{equation}

Decreasing $\lambda$ towards 0 would lessen the watermarking artifacts therefore improve imperceptibility but at the same time decrease bit accuracy. In contrast, higher value of $\lambda$ (up to the clamping upper bound of 1) would strengthen the residual term \textbf{r} and effectively increase bit accuracy at cost of image quality. \cref{fig:rescale} demonstrates this trade-off control. A reasonable range of $\lambda$ that satisfies both the PSNR and bit accuracy metrics is [0.75, 1.50]. We note that this trade-off can be controlled at sample level, meaning end-users can set their preferred trade-off for individual cover images.

\section{Qualitative evaluation}

\subsection{FFL effects}
\cref{fig:freq_ffl} shows the impact of FFL on the frequency spectrum of the watermarked images. Compared with TrustMark without FFL (\cref{fig:freq_ffl} middle), having this loss improves image quality across multiple frequency bands (note the brightness of the 4 regions at upper, lower, left and right of  \cref{fig:freq_ffl} right image).

\subsection{Post-processing layers}
We visualize several activation maps of the pre- and post-process layers in \cref{fig:pre}, along with the learned watermark and output image. The watermark is shown to located at the center of the image where salient objects are dominant and also probably because of random crop simulation during training. The watermark is shown visibly blended within the image content after pre-processing and perceptually invisible after the post-processing layers. 

\subsection{TrustMark on control images}

We visualize the watermarking artifacts by applying TrustMark on a set of control images with a fixed watermark in \cref{fig:eg_control}. The learned watermark is shown to be adaptive to the image content (most visible when comparing the third column with the first and second columns). Additionally, the watermark complexity increases along with the content (left to right), also the diagonal object edges tend to cause more artifacts than the horizontal and vertical edges. 

\begin{figure}[]
    \centering
    \includegraphics[width=1.0\linewidth]{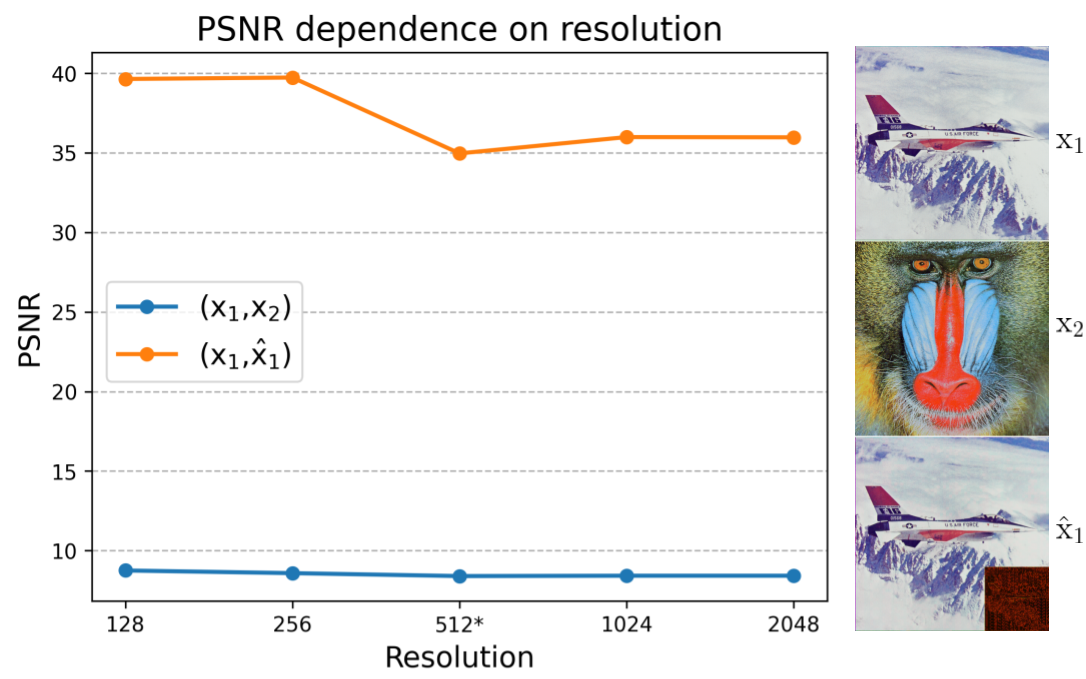}
    \caption{PSNR is dependent to the evaluating resolution and is more sensitive if the two images are similar (as in the case of watermarking). * refers to the original resolution of the evaluated images. $\hat{\mathrm{x}}_1$ inset: pixel residual scaled 20 times for visualization purpose.}
    \label{fig:psnr_dep}
\end{figure}

\section{Application demo}
We enclose a video demo demonstrating an use case of TrustMark for content provenance. Here, the watermark contains the hash key of the provenance manifest associated with the image. When uploaded to a social media platform (X), the provenance information can be recovered by using the extracted watermark to perform a key-value lookup in the provenance database. TrustMark works seamlessly with the C2PA open standards \cite{c2pa}, as shown in the demo. 

\end{appendices}
\end{document}